%% file: main.tex
\documentclass[lettersize,journal]{IEEEtran}

\usepackage{graphicx,color}
\usepackage{amsmath, amsthm, amsfonts, amssymb, amsbsy,nccmath}
\usepackage{mathtools}

\usepackage{algorithm}
\usepackage{enumerate}
\usepackage{lipsum}
\newtheorem{lemma}{Lemma}
\usepackage{textgreek}
\usepackage{textcomp}

\usepackage{algorithmic}

\usepackage[sort,compress]{cite}
\usepackage{epsfig}
\usepackage{epstopdf}
\usepackage{mathtools}
\usepackage{dsfont}
\usepackage{epstopdf}

\usepackage{sidecap, caption}
\usepackage{subcaption}
\theoremstyle{definition}
\newtheorem{definition}{Definition}

\theoremstyle{assumption}
\newtheorem{assumption}{Assumption}[section]

\theoremstyle{proposition}
\newtheorem{proposition}{Proposition}[section]
\theoremstyle{corollary}
\newtheorem{corollary}{Corollary}[section]

\usepackage[inline]{enumitem}   
\makeatletter
\newcommand{\inlineitem}[1][]{%
\ifnum\enit@type=\tw@
    {\descriptionlabel{#1}}
  \hspace{\labelsep}%
\else
  \ifnum\enit@type=\z@
       \refstepcounter{\@listctr}\fi
    \quad\@itemlabel\hspace{\labelsep}%
\fi}
\makeatother

\setlist[enumerate,1]{leftmargin=0.5cm}
\newcommand\norm[1]{\left\lVert#1\right\rVert}

\newtheorem{theorem}{Theorem}

\input{defines}

\makeatletter
\newcommand\fs@spaceruled{\def\@fs@cfont{\bfseries}\let\@fs@capt\floatc@ruled
  \def\@fs@pre{\vspace{0.5\baselineskip}\hrule height.8pt depth0pt \kern2pt}%
  \def\@fs@post{\kern1pt\hrule\relax}%
  \def\@fs@mid{\kern2pt\hrule\kern2pt}%
  \let\@fs@iftopcapt\iftrue}
\makeatother

\newcommand{\bit}{\begin{itemize}}
\newcommand{\eit}{\end{itemize}}

\newcommand{\mL}{\mathcal{L}}
\newcommand{\mwL}{\widehat{\mL}}

\newcommand{\bmY}{\boldsymbol{Y}}

\newcommand{\bpsi}{\boldsymbol{\psi}}

\newcommand{\bPhi}{\boldsymbol{\Phi}}

\renewcommand{\bmA}{{\boldsymbol A}}

\newcommand{\bmX}{{\boldsymbol{X}}}

\DeclarePairedDelimiter\abs{\lvert}{\rvert}%
\usepackage{stackengine}
\def\delequal{\mathrel{\ensurestackMath{\stackon[1pt]{=}{\scriptstyle\Delta}}}}

\newcommand{\bmuh}{\widehat{\bmu}}

\newcommand{\bmWh}{{\widehat{\bmW}}}
\usepackage{lipsum}

\makeatletter
\newcommand\longleftrightarrowfill@{%
  \arrowfill@\leftarrow\relbar\rightarrow}
\makeatother

\restylefloat{algorithm}
\usepackage{geometry}
\begin{document}

\title{ Neuro-Symbolic Causal Reasoning Meets Signaling Game \vspace{-1mm}for Emergent Semantic Communications\vspace{-1mm}}
\vspace{-0mm}\author{
\IEEEauthorblockN{\normalsize
Christo Kurisummoottil Thomas, \IEEEmembership{\normalsize Member, IEEE} and Walid Saad, \IEEEmembership{\normalsize Fellow, IEEE}
\vspace{0mm}}\\
\vspace{-1mm}
\thanks{Christo Kurisummoottil Thomas and Walid Saad are with the Wireless@VT, Bradley Department of Electrical and Computer Engineering, Virginia Tech, Arlington, VA, USA,(emails:\{christokt,walids\}@vt.edu). \\ \vspace{-0mm}
\indent A preliminary version of this work  appeared in IEEE GLOBECOM 2022 \cite{ThomasGC2022}. This research was supported by the Office of Naval Research (ONR) under
MURI grant N00014-19-1-2621.  }}
\maketitle
\vspace{-4mm}
\begin{abstract}\vspace{-0mm}
Semantic communication (SC) is an effective approach to communicate reliably with minimal data transfer while simultaneously providing seamless connectivity. In this paper, a novel emergent SC (ESC) framework is proposed. This ESC system is composed of two key components: A signaling game for emergent language design and a neuro-symbolic (NeSy) artificial intelligence (AI) approach for causal reasoning. In order to design the language, the signaling game is solved using an alternating maximization between the transmit and receive nodes utilities. The generalized Nash equilibrium is characterized, and it is shown that the resulting transmit and receive signaling strategies lead to a local equilibrium solution. As such, the emergent language not only creates an efficient (in physical bits transmitted) transmit vocabulary dependent on communication contexts but it also aids the reasoning process (and enables generalization to unseen scenarios) by splitting complex received messages into simpler reasoning tasks for the receiver. The causal description (symbolic component) at the transmitter is then modeled using the emerging AI framework of  generative flow networks (GFlowNets), whose parameters are optimized for higher semantic reliability. Using the reconstructed causal state, the receiver evaluates a set of logical formulas (symbolic part) to execute its task. This evaluation of logical formulas is done by combining GFlowNet, with the logical expressiveness of the symbolic structure, inspired from logical neural networks. The ESC system is also designed to enhance the novel semantic metrics of information, reliability, distortion and similarity that are designed using rigorous algebraic properties from category theory thereby generalizing the metrics beyond Shannon's notion of uncertainty. Simulation results confirm that the ESC system effectively communicates with reduced bits and achieves superior semantic reliability compared to conventional wireless systems and state-of-the-art SC systems lacking causal reasoning capabilities. Additionally, the overhead involved in language creation diminishes over time, validating the system's ability to generalize across multiple tasks.
\end{abstract}\vspace{-0mm}
\begin{IEEEkeywords}\vspace{-0mm}
 \small Semantic communications, neuro-symbolic AI, emergent language, signaling game, generative flow networks.
\end{IEEEkeywords}

\vspace{-3mm}
\section{Introduction}
\label{section_intro}
\vspace{-1mm}

\IEEEPARstart{F}{uture} wireless systems (e.g., 6G) must be more judicious in what they transmit to integrate high-rate, high-reliability, and time-critical wireless applications such as smart cities, industrial robotics, and the metaverse. Conventional wireless systems focus on reliably sending physical bits without emphasis on the \emph{semantic and effectiveness} layer, as pointed out by Shannon \cite{ShannonBSTJ1948}. Integrating the semantic and effectiveness aspects to create artificial intelligence (AI)-native wireless networks requires 
to only send information that is useful for the receiver. This is central premise of \emph{semantic communication (SC) systems} \cite{ChaccourArxiv2022,PopovskiJIIS2020,KountourisCommMag2021,XieTSP2021}. SC goals can be achieved by transitioning the transmit and receive nodes from being just blind devices toward brain-like devices capable of understanding and reasoning (using \emph{causal reasoning}) over the data and how it gets generated. 
  However, classical AI systems that rely purely on data statistics are not apropos for causal reasoning. Moreover, classical models cannot generalize to new data, and therefore, they cannot achieve the required reliability in real-world wireless scenarios.  

\vspace{-4mm}\subsection{Related Works}
\vspace{-1mm}

Despite the recent surge in works on SC \cite{XieTSP2021,SeoMehdiArxiv2022,LiuISIT2021,Mehdi2021, FarshbafanArxiv2022,FarshbafanICC2022}, most of them fail to address a few critical challenges. First, they mostly ignore the causal reasoning behind the data generation. Instead, they limit the discussion to the computation of meaningful attributes describing the observed data. However, this does not help achieve the benefits behind SC since the aforementioned methods are not built based on \emph{semantic awareness} about how the receiver uses the information. Second, the solutions of \cite{XieTSP2021,SeoMehdiArxiv2022,Mehdi2021,LiuISIT2021,FarshbafanArxiv2022} are not generalizable enough to situations not encountered during training and may require significant retraining efforts resulting in larger communication overheads. Here, although the authors of \cite{XieTSP2021} briefly mention the advantages of language models for reasoning and task generalization, their approach assumes prior knowledge of semantics. Their primary focus is optimizing the semantic and channel encoders using conventional information-theoretic metrics, which may not always minimize data transfer. 
Given the computing and memory constraints at the communicating nodes, it is desirable to have a language model between network nodes that evolves via an understanding of the node's mutual intentions and learns the semantics in a \emph{compositional} manner, i.e., combining knowledge of simpler concepts to
describe richer concepts. Such a model, called \emph{emergent language}, guided by causal reasoning helps generalize to unseen tasks without significant retraining effort and with possibly less training data. 
However, none of the existing SC frameworks \cite{XieTSP2021,SeoMehdiArxiv2022,LiuISIT2021,Mehdi2021, FarshbafanArxiv2022} is guided by the principles of causal reasoning and a compositional emergent language that could help reduce the AI training overheads and yet produce the SC system goals. { While the reasoning over the air approach of \cite{XiaoICC2022}, is intriguing, the authors utilize conventional reinforcement learning methods to infer implicit relations from transmitted data. This approach has a number of limitations, including poor sample efficiency of the AI algorithm and a lack of generalization to unseen data.}

\vspace{-4mm}\subsection{Contributions}
\vspace{-1mm}

In contrast to the prior art, the main contribution of this paper is a novel, fundamental design of the principles of \emph{emergent SC (ESC)} systems that include: (a) creating an \emph{emergent language} that learns the transmit and receive signaling strategies in a \emph{compositional} manner, and (b) instilling concrete reasoning capabilities (using novel AI methods) into the nodes enabling them to identify causal structure behind the data generation. {To instill reasoning, we use \emph{neuro-symbolic (NeSy) AI } to combine learning from experience (neural component) and reasoning from what has been learned (symbolic component). This enables the SC system to learn signaling strategies for complex tasks with fewer training data and communicate efficiently with fewer bits.} 
Key applications of this include goal oriented communications, mobile broadband and connected intelligence domains such as autonomous
driving, smart manufacturing, brain-computer interaction, flying vehicles, extended reality, and the digital twins \cite{ChristoJSAITArxiv2022}. In summary, our key contributions are the following:
\begin{itemize}
\vspace{-1mm}
\item To create an emergent language that is \emph{compositional} (helps generalization) and \emph{semantics aware} (for minimal transmission), we propose a novel formulation based on a two-player contextual signaling game (\emph{emergent language game}) for optimizing the speaker and listener signaling strategies.
\item To compute the game objective functions, we propose a novel semantic information measure derived from functors and enriched categories in category theory. This captures the information content conveyed by a transmit message across all possible worlds of logical interpretations in the semantic category, providing a more comprehensive and rigorous characterization compared to \cite{Carnap52}. 
\item We propose an alternating maximization (AM) based algorithm for computing the encoder and decoder distribution that converges to a \emph{local generalized Nash equilibrium (NE)}. Moreover, we theoretically prove that the optimal transmit strategy is partitioning the syntactic space into a \emph{Voronoi tessellation} and a Bayesian receive strategy to reconstruct the state description. This leads to a locally optimal solution for an emergent language that allow the speaker to transmit as minimum as possible and let the listener reconstruct more information.  
\item We exploit the potential of the recently proposed AI algorithm called generative flow networks (GFlowNet) \cite{BengioNeurIPS2021} for inducing causal reasoning at the nodes. Specifically, GFlowNet helps the speaker to compute the state description (causal reasoning about the data) which helps to have a smaller cardinality representation space for the emergent language. At the listener, reasoning via GFlowNet framework (a symbolic component at the listener) allows it to deduce the relevant logical conclusions from the reconstructed data to complete its task with the required semantic reliability. 
For any previously unseen, more arduous task, a chain of thought type reasoning (inspired by \cite{WeiChainReasonArxiv2022}) implemented by GFlowNet and facilitated by the language compositionality allows the listener to generalize the capabilities without any retraining. The resulting NeSy AI approach (\cite{GarcezArxiv2020}) saves significant computational efforts and communication overheads arising from retraining.
\item We show that ESC systems can encode data with minimal bits compared to a classical system that does not exploit causal reasoning and where the speaker is unaware of the semantic effectiveness at the listener. Furthermore, we prove that ESC systems can achieve higher semantic reliability (with fewer physical bits) by reasoning over the created language and exploiting the contextual information than by treating the data at the bit level as in Shannon's information theory. 
\end{itemize}\vspace{-1mm}

Simulation results show that the proposed game-based emergent language converges much faster compared to state of the art algorithm \cite{Mehdi2021} that relies upon variational inference and {does not accurately maximize SC goals}. Moreover, the overhead associated with our proposed emergent language training duration exponentially decreases over time with task variations, signifying its task agnostic nature. Our results also show that the expected semantic representation length for an ESC system gets significantly reduced compared to a classical system by considering contextual information and causal reasoning. This justifies our ESC system principle of ``transmitting less and doing more with the data."  We also show that our ESC system (with trained GFlowNet as a causal component) can achieve significantly better semantic reliability than the state-of-the-art, including classical AI-based methods.

The rest of this paper is organized as follows. In Section II, we present the proposed ESC system model. Section III describes the proposed two-player contextual signaling game. Section IV introduces the reasoning network formulation using GFlowNet. Section V provides simulation results. Finally, conclusions are drawn in Section VI. We use a standard approach for our notation\footnote{ \scriptsize\emph{Notation:} Lower-case letter $a$ is a scalar, lower-case boldface letter $\bma$ is a vector, and boldface upper-case letter $\bmA$ is a matrix. A set (either discrete or continuous entries) or a topological space is represented using Calligraphic font $\mX$. $\mR^M$ represents an $M-$dimensional vector whose entries belong to real numbers $\mR$. $[a,b]$ represents the real number range between $a$ and $b$. $D_{KL} (p||q)$ represents the KL divergence (KLD) between $p$ and $q$. $\mZ$ denotes the set of integers. $\land$ represents the logical AND operator $\implies$ is the logical implied by.
} hereinafter.

\vspace{-3mm}
\section{Proposed ESC System Model}
\label{Section_SCModel}
\vspace{-1mm}

\begin{figure*}[t]\vspace{-1mm}
\centerline{\includegraphics[width=6.6in,height=2.4in]{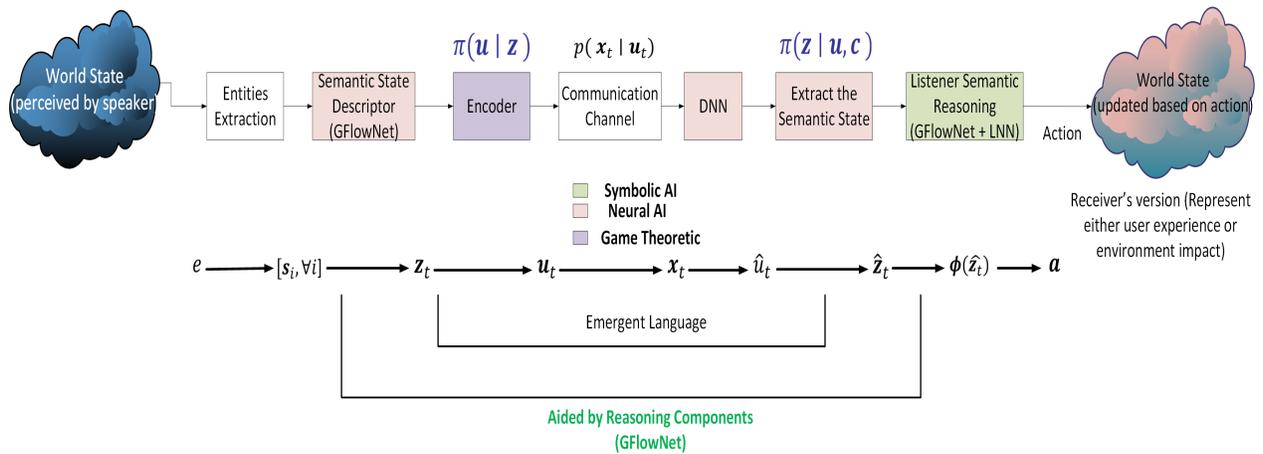}} 
\vspace{-3mm}\caption{\scriptsize Overview of proposed ESC model. Our focus is on 1) Learning of the state description, which is grounded on the real logic and formed based on the semantic language rules. 2) Encoder/decoder functions that translate the states to optimal physical messages with the objective of transmitting less but with semantic reliability close to $1$. Objectives are achieved via the tools of NeSy AI.}
\label{Fig_SCModel}
\vspace{-2mm}
\end{figure*}
Consider a speaker who observes a sequence of events that capture
the state of the environment, and each event $e \in \mathcal{E}$. The speaker intends to send a concise description of the event (in other words, a causal description which represents the \emph{semantics} of information) to a
listener who wishes to complete a set of sequential tasks drawn from a probability distribution $p(\tau)$. From the inferred semantics, the listener further executes an action represented by the probability distribution $p(
\bma \mid \bPhi)$ and $\bma \in \mathcal{A}$ ($\subset \mathcal{R}$, a real value). $\bPhi$ is the set of logical conclusions that the listener is interested in evaluating and represents the {listener} semantics. The effectiveness of the actions on the listener's environment represents the \emph{semantic effectiveness} aspect. We assume that the impact of the listener's action gets reflected in the subsequent events observed by the speaker. This means that communication occurs in a setting in which the listener's environment overlaps partially with that of speaker's. However, herein, at the start of communication, the speaker is unaware of the listener's intentions and its logical interpretations of the received semantics. Hence, apart from the causal reasoning components at the nodes, a communicating language (denoted the emergent language) has to be constructed if the SC system is to achieve the benefits outlined in Section~\ref{section_intro}. The resulting ESC system framework is shown in Fig.~\ref{Fig_SCModel}. Here, the speaker first extract the different features from $e$. These features and their properties are called \emph{entities}. {We represent the set of $N$ entities as the set of vectors $\{\bms_0,\cdots,\bms_{N-1}\}$, where each entry $\bms_i \in \mathcal{R}^M$ (grounded to real logic).} 
Our approach is versatile and applicable to diverse data types as the causal state descriptor, listener semantic reasoning component, and emergent language are designed to be general, enabling adaptation to different applications and domains without constraints on specific data modalities.

\vspace{-4mm}
\subsection{Causal Reasoning Model}
\label{eq_CRM}
\vspace{-1mm}

Next, our objective is to infer the hidden relations among the entities and compute the causal sequence that best explains the event observed. This component is called the \emph{semantic state descriptor} (a neural component) in Fig.~\ref{Fig_SCModel}. The semantic state description is defined as the sequence of entities and their relations, $\small {\widetilde{\bmz} \!=\! \left(\bms_0,\bmr_1,\cdots,\bmr_{N-1},\bms_{N-1}\right)}$, of dimension {$M(2N-1)\!\times\!1$}. Here, the vector $\bmr_i \in \mathcal{R}^M$ denotes the relation between entities $\bms_i$ and $\bms_{i+1}$. One way to represent $\bmr_i$ is as a linear relation, $\bms_{i+1} \!\approx \!\bmr_i\! + \!\bms_{i}$, where $\bmr_i = -\ln p(\bms_{i+1}\mid \bms_i)$.
If $\bms_i$ and $\bms_{i+1}$ are not causally related $p(\bms_{i+1}\mid \bms_i)=0$. Hence, when the entities are far apart in the grounded domain $\mathcal{R}^M$, then $\bmr_i$ has entries $\infty$. When the  $p(\bms_{i+1}\mid \bms_i) \approx 1$, then the entities are very close to each other in $\mathcal{R}^M$.
{A modular approach here to compute \emph{the cause and effect relations} is to utilize the structural causal model (SCM) described in \cite{BareinboimACM2020} and \cite{ScholkopfPIEEE2021}. The SCM framework considers a group of observables, represented by the vertices of a directed acyclic graph (DAG), denoted as $\bms_i$. It is assumed that each observable is the result of an assignment of the form:
$\bms_i = f_i(\textrm{pa}(\bms_i),\bmv_i), \forall i $
where $f_i$ is a deterministic function that relies on the parents of $\bms_i$, $\textrm{pa}(\bms_i)$ in the graph, and an unknown random vector $\bmv_i$ of dimension $M\times 1$. The noise term $\bmv_i$ ensures that the SCM assignment above can represent a general conditional distribution $p(\bms_i\mid \textrm{pa}(\bms_i))$. Following this, we propose to learn the causal DAG \cite{ScholkopfPIEEE2021} by investigating the posterior distribution of entities as: }
\vspace{-1mm}\beq  \vspace{-1mm}
 \begin{array}{l}\vspace{-1mm}
\mP_S:  p(\bms_0,\bms_1,\cdots,\bms_{N-1}\mid e)= \prod\limits_{i} p(\bms_i \mid \textrm{pa}(\bms_i)).   \end{array}
 \label{eq_DAG_causality}
 \vspace{-1mm}\eeq
In a DAG, there is an edge from node $\bms_i$ to $\bms_{i+1}$ when they are causally related per \eqref{eq_DAG_causality}. The $\bmr_i$ learned when computing \eqref{eq_DAG_causality} does not need to be transmitted, rather only a truncated version of $\widetilde{\bmz}$ (by removing $\bmr_i$), that is $\small {{\bmz} \!=\! \left(\bms_0,\bms_1,\cdots,\bms_{N-1}\right)}$, which is from hereon called the \emph{state description}. This assumes that, given $\bmz$, the listener can learn $\bmr_i$ by computing the posterior \eqref{eq_DAG_causality} at its end and hence reduce transmission. 
GFlowNet is chosen here since it incorporates causality and can also learn the exact (or close to optimal) posterior distribution under different variations in the task distribution (in short, a multi-modal distribution compared to variational techniques \cite{HoffmanJMLR2013}). {Furthermore, the interventions can be incorporated in the proposed causality model by considering different probability distributions for the various factors in \eqref{eq_DAG_causality}. This can be analytically written as (superscript $e_i$ denotes a particular environment)
 \vspace{-1mm}\beq  \vspace{-0mm}
 \begin{array}{l}\vspace{-1mm}
 p^{(e_i)}(\bms_0,\cdots,\bms_{N-1}\mid e)\\[1mm] = \prod\limits_{i \in \mI \backslash \mI_G} p(\bms_i \mid \textrm{pa}(\bms_i))\prod\limits_{j\in \mI_G} p(\bms_i \mid \textrm{pa}(\bms_i)),   \end{array}
 \label{eq_DAG_causality_I}
 \vspace{-0mm}\eeq
where $\mI$ is the set of all indices $i$ and $\mI_G$ represents the set of interventions performed on the DAG (similar concepts to implement interventions for SCM appear in \cite{AnsonArxiv2022}).}

After signal reception, the listener intends to derive certain logical conclusions (called \emph{logical formulas}) from the {reconstructed state description $\bmzh_t$} to execute a task or to perform other system purposes. The capability to make logical conclusions from the received data is vital to enable human-like reasoning (i.e., cognition/common sense) that can enable them for automated decisions or efficient data recovery in future intelligent wireless systems. This is the \emph{listener semantic reasoning} (symbolic) component in Fig.~\ref{Fig_SCModel}. 
The listener has a set 
 $\bPhi$ of $L$ logical formulas that are evaluated using $\bmzh_t$ as
\vspace{-1mm}\beq
\vspace{-1mm}\mP_L: \mbox{Compute}\,\,\, p(\bphi_i \mid \bmzh),\, \,\forall \bphi_i \in \bPhi. \,\,\, 
\label{eq_list_semreasoning}
\vspace{-1mm}\eeq
{We define the logical formula $\bphi_i$ using the connectives $\land$ and $\implies$ as follows:
\vspace{-1mm}\beq
\bphi_i \delequal \left(l_1^{(i)}(\bmzh) \land l_2^{(i)}(\bmzh) \cdots \land l_{D-1}^{(i)}(\bmzh) \implies l_{D}^{(i)}(\bmzh)\right).
\label{eq_phi_i}\vspace{-1mm}
\eeq
$\bphi_i$ here evaluates to either true or false (represented probabilistically as a real number between $[0,1]$). Each term $l_j^{(i)}(\bmzh)$ represent a probability value and hence in the range $\left[0,1\right]$. The logical formulas in \eqref{eq_list_semreasoning} can be efficiently computed using a combination of GFlowNet (that implements $l_j^{(i)}(\bmzh)$) \cite{ZhangBengio2022} and logical connectives inspired from logical neural networks (LNN) \cite{RiegelLNN2020} \footnote{\scriptsize In contrast to LNN \cite{RiegelLNN2020}, which presupposes given logical symbols and does not extract causal structures, the proposed NeSy AI method, which combines GFlowNet and NN-based logical connectives (inspired by LNN), can offer a more comprehensive representation of causal aspects. This is because NeSy AI utilizes transformers that enable parallel processing of entities, whereas LNN uses recurrent neural networks (RNN) that process entities sequentially.}.} 

\vspace{-3mm}
\subsection{Emergent Language Model}
\vspace{-1mm}

We have thus far defined the {causal state description} that the speaker derives from the event and the listener extracts from the received information. 
Further, we need to map $\bmz$ ($\in \mathcal{W}$, the \emph{syntactic space}) to the encoded representation $\bmu$ ($\in \mathcal{U}$, a finite vocabulary) that gets transmitted. 
For an ESC system in which the intentions of both nodes are oblivious to one another, one approach for the speaker to learn semantic awareness is via an interactive conversation. Hence, we propose a novel two-player contextual signaling game to learn the speaker transmit distribution $\pi_s=\pi(\bmu\mid \bmz)$, the listener inference distribution $\pi_l=\pi(\bmz\mid \bmu,c)\pi(c\mid \bmu)$, where the {context $c$ belongs to a finite set, $\mathcal{C} \subset \mathcal{Z}$}.  \emph{Communication context} here represents the set of all possible {state descriptions} and the valid logical interpretations defined for a particular scenario corresponding to a task. {The communication context, denoted by $c \in \mathcal{C}$, captures the richness of the syntactic space in our proposed approach. Specifically, the set of numerical features $\bms_i$ represents a causal state description, where $\bms_i \in \mS_i$ for each context $c_i$ of a given task $\tau$. There may be possible overlap between different $\mS_i$.  $c_i$ can be represented as an integer value, indicating which $\mS_i$ corresponds to that particular context.} Given the syntactic space $\mathcal{W}$, context set $\mathcal{C}$, the representation space $\mathcal{U}$, 
and the set of all $\pi_{s}$ and $\pi_{l}$ which are $\mathcal{M}_{s}$ and $\mathcal{M}_{l}$, respectively, 
we define the \emph{signaling game} as the tuple $(\mathcal{C},\mathcal{W},\mathcal{U},\mathcal{M}_{s},\mathcal{M}_{l})$. 
The use of a game-theoretic approach here is apropos because, at equilibrium, the transmitter will obtain a semantically aware encoder that transmits minimal information such that the listener recovers the state description closest semantically to the transmitted version. Using encoded representations for each entity in an event and letting the listener extract semantics from them is not communication efficient due to the large number of bits needed to represent all entities. 

Having outlined a general SC paradigm for several future wireless scenarios, we next look at how we design the emergent language and causal reasoning processes that help us achieve the ESC gains.
\begin{figure*}[t]\vspace{-2mm}
\begin{subfigure}{0.52\textwidth} 
\centerline{\includegraphics[width=3.9in,height=1.3in]{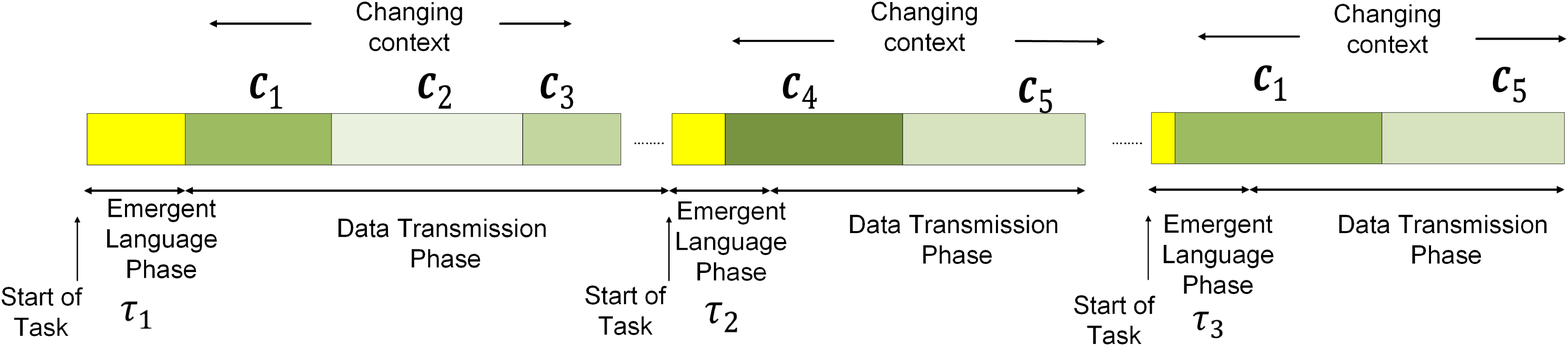}} 
\vspace{-4mm}\caption{}
\vspace{-2mm}
\label{Fig_timesplit}
\end{subfigure}
\begin{subfigure}{0.48\textwidth}
\centerline{\includegraphics[width=3.2in,height=1.3in]{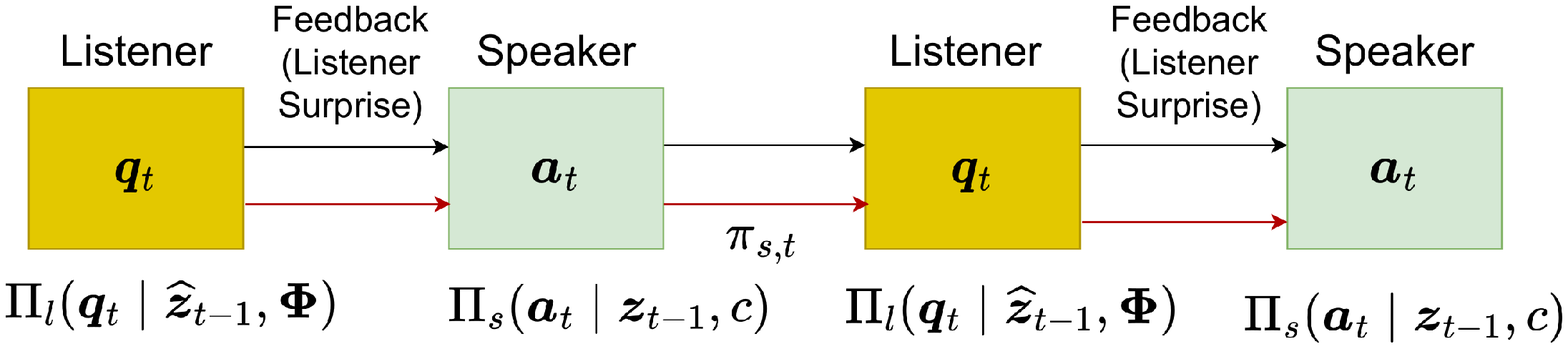}}
\vspace{-4mm}\caption{}
\vspace{-2mm}
\label{Fig_EmLang}
\end{subfigure}
\vspace{-1mm}\caption{\scriptsize (a) Proposed ESC system time split between emergent language and data transmission phases. b) Illustration of emergent language.}
\label{Fig_TimeLine}\vspace{-3mm}
\end{figure*}

\vspace{-1mm}\section{Proposed Signaling Games For Semantic Emergent Language Reasoning}
\label{EmergentLanguage}
\vspace{-1mm}

Given our ESC model, Fig.~\ref{Fig_timesplit} shows how the emergent language training and data transmission evolves. In the emergent language phase, given the semantic state description $\bmz$, the objective is to derive an optimized representation space $\mathcal{U}$, computed as the probability distribution $\pi_s$. An optimized $\mathcal{U}$ allows the speaker to reduce its transmission and the listener to extract maximum semantic information to execute its task effectively during data transmission phase. In order to ensure that the nodes can efficiently communicate with desired reliability on tasks unseen during training (defined as \emph{generalizability}), the emergent language must be constructed on the principle of \emph{compositionality}. Compositionality can be defined as reasoning on complex state descriptions by combining conclusions derived from smaller sub-parts of the state description. 
While our emergent language may have an initial overhead, its duration will reduce over time, as shown in Fig.~\ref{Fig_timesplit}. This is because the learned emergent language becomes generalizable and task agnostic over time. We cast this problem as a two-player contextual signaling game, called \emph{emergent language game}. 

\vspace{-3mm}\subsection{Two-player Contextual Signaling Game Model for Emergent Language}
\label{section_EmLanguage}
\vspace{-1mm}

Motivated from the visual question and answering (Q/A) games in AI \cite{AndreasCVPR2016}, we define a
signaling game \cite{FudenbergCambridge1991} between the speaker and listener as follows. The language, as shown in Fig.~\ref{Fig_EmLang}, evolves via a series of questions $\bmq_t\in \!\mathcal{W}$ (question) by the listener and answers provided by the speaker, $\bma_t\!\in\! \mathcal{W}$ (answer), which help the listener reconstruct the description of the perceived event or partial set of features to achieve its intent. $\bmq_t$ and $\bma_t$ are vectors of dimensions $M\times 1$. {The Q/A-scheme proceeds over multiple communication rounds  wherein the answer $\bma_t$ is encoded using the transmit distribution $\pi_{s,t}$ which is learned. Each question $\bmq_t$ transmitted by the listener belongs to a small finite set. Hence, it can be encoded using fewer bits (for example, using an $M-$QAM, where $M < 100$, assuming number of questions are as low).} We next define the question and answer concepts.
\vspace{-1mm}\begin{definition} 
\vspace{-1mm}
 {We define an \emph{question} as one that involves just one entity and expects either an entity or a Boolean variable as an answer (which is the \emph{answer}).}
 \vspace{-1mm}\end{definition}
 Each answer represents a simpler {state description} within a context $c$ and corresponds to a task $\tau$. Hence, the listener adapts to the ESC system by learning simpler concepts.  $\mathcal{W}$ represents the syntactic space that defines the set of possible entities used to describe the state description. 
\vspace{-1mm}\begin{assumption}  The emergent language is created under the following assumptions: 1) The syntactic space $\mathcal{W}$ is assumed to be known at both nodes before the emergent language phase. However, the listener is unaware of the event dynamics (the distribution that generates $\bmz_t$), and 2) only the listener is aware of the semantics stemming from any state $\bmz_t$. 
\vspace{-1mm}\end{assumption}
The above assumptions outline the known variables at both nodes before the start of communication. 
At the emergence of the language, the speaker has an optimized $\pi_s^*$ (with $\bmu \in \mathcal{U}$) and the listener can reliably reconstruct the state description using $\pi_l^*$ to execute the task at hand. The idea here is that 
{listener} strategies can infer the context to construe the very same signal differently given
different communication contexts. This allows the listener to disambiguate signals that are ambiguously
used by the sender. 
At any communication round $t$, listener's state is $\bmzh_t = (\tau,\bmq_0,\bma_0,\cdots,\bmq_{t-1},\bma_{t-1},\widehat{c}_{t-1},\bmq_t)$. $\widehat{c}_{t-1}$ represent the inference about the context $c$ at time $t-1$. 
Similarly, the speaker has the state, $\bmz_t = (e,c,\bmq_0,\bma_0,\cdots,\bmq_{t-1},\bma_{t-1},\bmq_t)$, after having observed the question $\bmq_t$ from the listener. The listener samples a question based on a stochastic policy  $\Pi_l(\bmq_t\mid \bmzh_{t-1},\bPhi)$. Question $\bmq_t$ captures the logical formula the listener wants to evaluate from the set $\bPhi$. {The set of these questions is predetermined, belongs to a fixed set, and is specific to the set of logical formulas the listener wants to evaluate. Meanwhile, the speaker samples the answer from the stochastic policy $\Pi_s(\bma_t\mid \bmz_t,\bmc)$. Here, the answer is not predetermined, and it depends on the sampled event}. $\Pi_s$ is the event distribution and represents a method of generating the next {entity} given the current state sequence $\bmz_t$ and the context $c$. The state sequence $\bmz_t$ depends on the listener's intent ($\bPhi$) and the previous sequence of answers from the speaker. These stochastic policies $\Pi_l$ and $\Pi_s$ are assumed to be known on the listener and speaker sides, respectively. The syntactic space $\mathcal{W}$ can be infinitely large depending on the context, but the ESC system aims to transmit less and do more with small data. To achieve this, the speaker restricts the encoded representation space ($\mathcal{U}$) to be finite, and the optimized vocabulary implicitly determines the number of communication resources allocated to a listener node. Semantic information measures are preferred over classical mutual information metrics to accurately represent semantic effectiveness. 
We define $S_s(\bmz_t;\bmu_t\mid \pi_{s,t},[\bmz_{t-1}],c)$ as the semantic information conveyed by the speaker about the state $\bmz_t$ given $\bmu_t$, under a fixed $\pi_{s,t}$, $c$ and given the past states denoted by $[\bmz_{t-1}]$. Similarly, on the { listener} side, the amount of information acquired by the listener about state $\bmzh_t$ for fixed policies and given the previous extracted states $[\bmzh_{t-1}]$ is $S_l(\bmzh_t;\bmu_t\mid \pi_{s,t},\pi_{l,t},[\bmzh_{t-1}])$. We next formally define $S_s$ and $S_l$. 

\vspace{-3mm}
\subsection{Semantic Information Measure}
\vspace{-1mm}

One approach to capture the semantic information is by using the propositional logic over sets introduced by \cite{Carnap52}. However, this approach does not generalize to reasoning-based emergent language because the set-theoretic approach does not have the compositionality properties which we discuss below. Our proposal is to utilize a category theory based method for representing semantic information that is more comprehensive than previous approaches in the field. 
\vspace{-1mm}\begin{definition}
\vspace{-1mm} We define a \emph{deductive logic} as the following conjunction of {relations} (if involving multiple entities), $r^{(N)} = (r_1 \land r_2 \cdots \land r_N)$.
Here, $r^{(N)}$ holds iff all the {relations} $r_i$ are true. 
\vspace{-0mm}\end{definition}
\begin{definition}\label{def_chain_like_rule}\vspace{-2mm}
We define \emph{chain-like rules} as follows, 
$(e_0,r^{(N)},e_N) \!=\!(e_0,r_1,e_1) \land (e_1,r_2,e_2) \land \cdots \land (e_{N-1},r_N,e_N).$
The chain rule will be essential for proving analytical or logical theorems using the emergent language. This follows also from the associative property of the relation triplets. 
\vspace{-1mm}\end{definition}
Hence, if the speaker or listener is to derive logical conclusions, defined as $\bphi_i$ in \eqref{eq_list_semreasoning}, from the existing or learned relations between the entities, the above properties hold the key. To this end, category theory has excellent algebraic structural properties that allows the grouping of all the syntactic and semantic objects part of the language. Preliminary definitions required for category theory terminologies used herein, such as copresheaves, functor, and enriched categories, are provided in Appendix~\ref{app_catTheory}. Next, we derive here a novel interpretation of the semantic representation of the meaning conveyed by the state $\bmz_t$ using category theory \cite{FongArxiv2018}. These novel representations of semantics lead us to define the semantic notion of information. The following definition constructs a category whose objects belong to $\mathcal{W}$.
\vspace{-1mm}\begin{definition}
\vspace{-1mm}Any entity, represented by $\bma_t$ or $\bmq_t$ or expressions part of $\mathcal{W}$ form objects in a category $\mL$, called the \emph{syntax category}. Expressions are defined as molecular entities that are a combination of multiple entities, e.g., the state $\bmz_t$. A \emph{morphism} (a directed relation between the objects in any category) from object $\bmx$ to object $\bmy$ in $\mL$ is formed when expression $\bmy$ extends from the expression $\bmx$.
\vspace{-1mm}\end{definition}

Moreover, category $\mL$ and the morphisms can be represented as a category enriched in $[0,1]$. This means that we can represent the morphism, $\bmx \rightarrow \bmy$ using a probability distribution, $\pi(\bmy\mid \bmx) \in [0,1],$ that captures the probability that expression $\bmx$ extends to $\bmy$. However, this category $\mL$ only represents the language's compositional part (or the syntax part) under consideration. Ideally, we should be able to encode ``what goes with what" (\emph{syntactic part}) together with the plausible, logical conclusions one can derive from the expressions (\emph{semantic part}). Further below, we show that \emph{semantic category} $\mwL$ can be represented  as a transformation from the syntax category, represented as, $F:\mL \rightarrow \mwL$. 
\vspace{-1mm}
\begin{lemma}
\vspace{-1mm}
\label{lemma_semanticcat}
The semantic category $\mwL=[0,1]^L$ within the context $c$ is an $[0,1]$ enriched category of representable copresheaves from the syntax category $\mL$. 
\vspace{-1mm}
\end{lemma}
\begin{IEEEproof}
    \vspace{-0mm}
The proof is given in Appendix~\ref{appendix_proof_lemma_semcat}.
\end{IEEEproof}
In Lemma~\ref{lemma_semanticcat}, copresheaves of any object $\bmx \in \mL,$ represent all the possible morphisms (expressions) from $\bmx$, within context $c$. 
The usefulness of the Lemma is two-fold. Firstly, at the listener, it helps to reason on any logical formula $\bphi_i$ (representing objects in the semantic category) by splitting a complex state description into smaller concepts and evaluating a set of smaller logical formulas $\bphi_j$. Further, evaluate $\bphi_i$ using the chain-like rule defined previously with $\bphi_j$'s as the components. This is only possible if we show that the semantic space is also a category. Secondly, leveraging Lemma~\ref{lemma_semanticcat}, we provide a novel definition of the semantic information conveyed by any expression $\bmx \in \mL$ as follows. 
\vspace{-1mm}\begin{lemma}
\vspace{-1mm}\label{theorem_semInf}
The \emph{semantic information} conveyed by any expression $\bmx\in \mL$ can be written as the average across the information carried by the representable copresheaf that follow from $\bmx$.  Further, mathematically, we can write the semantic information conveyed by $\bmx$ as
{\vspace{-1mm}
\beq
\vspace{-1mm}
S(\bmx\mid c) = \sum\limits_{\bmy \in \mH^\bmx} \pi(\bmy\mid \bmx,c)  \log \frac{\pi(\bmy\mid \bmx,c) }{\pi(\bmy)},
\eeq
such that the support of $\bmy$ is defined by the representable copresheaves $\mH^\bmx$.}
\vspace{-1mm}
\label{eq_SemInfMeasure_x_c_d}
\end{lemma}
\vspace{-1mm}
\begin{IEEEproof}\vspace{-1mm}
Lemma~\ref{lemma_semanticcat} provides a way to categorize the possible logical conclusions that entail from any $\bmx$. The resulting distributional representation provides a rigorous way to formulate the semantic information conveyed by any expression $\bmx$ about $\bmy$, as follows. Inspired by \cite{Carnap52,LindleyAMS1956}, we write 
\vspace{-1mm}
\beq
\vspace{-1mm} 
S(\bmy;\bmx\mid c) = \begin{cases}\log \frac{\pi(\bmy\mid \bmx,c) }{\pi(\bmy)},\,\,\, &\textrm{if}\,\, \bmy \geq \bmx \\
0,\,\,\, &\textrm{otherwise}.
\end{cases}
\label{eq_sem_y_x}
\vspace{-1mm}
\eeq
Further, we can obtain the semantic information captured by $\bmx$ as the average of the semantic information conveyed by possible worlds that contain $\bmx$ which is (4). 
\end{IEEEproof}
Intuitively, \eqref{eq_sem_y_x} represents the degree of confidence in a possible logical entailment $\bmy$ after having observed $\bmx$. {This means that if $\pi(\bmy\mid \bmx,c) = \pi(\bmy)$, then it means $\bmy$ does not entail from $\bmx$. Hence, the corresponding $S(\bmy;\bmx\mid c) = 0$; otherwise, it will be a quantity greater than $0$, representing non-zero semantic information. }Lemma~\ref{theorem_semInf} provides a way to compute the semantic information conveyed by $\bmz_t$ as
$S_s\left(\bmz_t\mid \pi_{s,t} [\bmz_{t-1}], c\right) \!=\!  \sum\limits_{\bmy\in \mH^{\bmz_t}}\pi(\bmy\mid [\bmz_{t}],c)\log\frac{\pi(\bmy\mid [\bmz_{t}],c)}{\pi(\bmy\mid [\bmz_{t-1}],c)}.$
 Further, we average across $\bmu_t$ to arrive at the average semantic information conveyed by $\bmz_t$ for a particular policy $\pi_{s,t}$ and context $c$ as in \eqref{eq_speakersemInfo}.
\vspace{-0mm}
\begin{figure*}\beq
\begin{aligned}
 \mathbb{E}_{\bmu_t} \left[S_s(\bmz_t;\bmu_t\mid \pi_{s,t},[\bmz_{t-1}],c)\right]    &=\sum\limits_{\bmz_t}\pi\left(\bmz_t\mid c,[\bmz_{t-1}]\right) \underbrace{\left[\sum\limits_{\bmu_t}\pi_{s,t}\log\frac{\pi_{s,t}}{\pi(\bmu_t)}\right]}_{\mbox{ \parbox{4cm}{avg. number of bits/state description}}} \underbrace{\vphantom{\sum\limits_{\bmu_t}}S_s(\bmz_t\mid [\bmz_{t-1}],c)}_{\mbox{semantics}},\!\! \\
 \mbox{where},\,\, \,\,\,\,\,\,\,\,\,\,\,\,\,\,\,\,\,\,\,\,\,\,\pi(\bmu_t)  &= \sum\limits_{\bmz_t}\pi_{s,t}\pi(\bmz_t\mid c,[\bmz_{t-1}]),\,\,\mbox{and}\,\, \pi_{s,t} = \pi(\bmu_t\mid [\bmz_t]).
\end{aligned}
\label{eq_speakersemInfo}
\vspace{-1mm}\eeq
\end{figure*}
An appropriate unit to represent $\frac{S_s(\bmz_t;\bmu_t\mid \pi_{s,t},[\bmz_{t-1}],c)}{\left[\sum\limits_{\bmu_t}\pi_{s,t}\log\frac{\pi_{s,t}}{\pi(\bmu_t)}\right]}$ will be \emph{average semantics/s/Hz}.
Next, we look at the average semantic information captured at the listener side, given that the actual transmitted state was $\bmz_t$ and for a specific listener policy $\pi_{l,t} = \pi(c,\bmzh_t\mid\bmu_t,[\bmzh_{t-1}])$. For this purpose, we first define the following corollary on \emph{semantic similarity} that follows directly from the copresheaves definition.
\vspace{-0mm}\begin{corollary}\vspace{-1mm}
\label{label_corollary_simMetric}
The similarity metric $Z_{xy}$ between two objects, $\bmx$ and $\bmy$,  can be defined as the intersection of the representable copresheaf between them.
\beq
\vspace{-1mm} \footnotesize Z_{xy} = \sqrt{\frac{\sum\limits_{\bmz\in \mH^{\bmx} \cap \mH^{\bmy}}h^{\bmx}(\bmz)}{\sum\limits_{\bmz\in \mH^x }h^{\bmx}(\bmz)}\frac{\sum\limits_{\bmz\in \mH^{\bmx} \cap \mH^{\bmy}}h^{\bmy}(\bmz)}{\sum\limits_{\bmz\in \mH^{\bmy} }h^{\bmy}(\bmz)}}.\vspace{-0mm}
\vspace{-1mm}\eeq
Whenever, $\mH^{\bmx} = \mH^{\bmy}$, then $Z_{ij}=1$ and when $\mH^{\bmx} \cap \mH^{\bmy} = \emptyset$, $Z_{ij} = 0$.
\vspace{-1mm}\end{corollary}
Leveraging corollary~\ref{label_corollary_simMetric}, we obtain the average semantic information extracted by listener as in \eqref{eq_semInfo_listener}.
\begin{figure*}
\vspace{-1mm}\beq
\vspace{-3mm}\begin{array}{l}
\mathbb{E}_{\bmu_t}\left[S_l(\widehat{\bmz}_t;\bmu_t\mid \bmz_t,[\bmzh_{t-1}],c,\pi_{l,t}) \right]   = \sum\limits_{\bmu_t}\pi\left(\bmu_t\mid [\bmz_t]\right)S_s({\bmzh}_t\mid [\bmzh_{t-1}],c)\left[\sum\limits_{\widehat{\bmz}_t,\widehat{c}}\pi_{l,t}\log\frac{\pi_{l,t}}{\pi(\widehat{\bmz}_t\mid c,[\bmzh_{t-1}])}Z_{\widehat{z}_tz_t}\right].
\end{array}\vspace{1mm}
\label{eq_semInfo_listener}
\eeq
\end{figure*}
In \eqref{eq_semInfo_listener}, $\small [\sum\limits_{\widehat{\bmz}_t,c}\pi_{l,t}\log\frac{\pi_{l,t}}{\pi(\widehat{\bmz}_t\mid c,[\bmzh_{t-1}])}Z_{\widehat{z}_tz_t}]$ represents the probability that $\bmz_t\!=\!\widehat{\bmz}_t$, semantically. We now have the necessary ingredients to define the two-player signaling game NE problem.

\vspace{-3mm}\subsection{Signaling Game Model and Generalized Nash Equilibrium Problem}
\vspace{-1mm}

Using the metrics above, we further formulate the solutions of the signaling game defined using the tuple $(\mathcal{C},\mathcal{W},\mathcal{U},\mathcal{M}_{s},\mathcal{M}_{l})$ as a \emph{Bayesian generalized NE}\cite{FacchineiAOR2010,FudenbergCambridge1991} (generalized due to the constraints part of the problem formulation below). The NE strategies are such that for all $\bmz_t \!\in\! \mathcal{W}$ and $c \!\in \!\mathcal{C}$
\begin{subequations}
\vspace{-1mm}\begin{align}
\pi_{s,t}^* \, \in & \,\arg\max\limits_{{\pi}_{s,t}} -\mathbb{E}_{\bmu_t}\left[S_s(\bmz_t;\bmu_t\mid {\pi}_{s,t},[\bmz_{t-1}],c)\right], \, \label{eqn_NE:1}\\  \vspace{-1mm}&\,\,\textrm{s.t.} \, \, \, \, \, \, \, \, \, \, \, \, \, \, \, \, \, \, \, \mathbb{E} [V({\pi}_{s,t},\pi_{l,t})] \leq D \label{eq_NE_constraint}\\ \vspace{-0mm}
\pi_{l,t} ^* \,\in & \,\arg\max\limits_{\pi_{l,t}} \mathbb{E}_{\bmu_t}\left[S_l(\bmzh_t;\bmu_t\mid {\pi}_{l,t},\pi_{s,t}^*,[\bmzh_{t-1}])\right],\label{eqn_NE:2}
\vspace{-1mm}\end{align}
\vspace{-0mm}\label{eq_NashEq_formulation}
\vspace{-0mm}\end{subequations}
\!\!where $V(\pi_{s,t},\pi_{l,t}) \!= \!c(\bmu_t) - \log \pi_{l,t},$
and the expectation of $V$ is over the policies $\pi_{s,t},\pi_{l,t}$. $c(\bmu_t)$ represents the cost of producing $\bmu_t$ and $-\!\log \pi_{l,t}$ represents the listener's surprisal about inferring $\bmz_t$ given $\bmu_t$. The equation~\eqref{eqn_NE:1} states that the strategy space $\mathcal{M}_{s,t}$ consists of signals $\bmu_t \!\in \!\mathcal{U}$ that minimizes the expected semantic information at speaker's end. Constraint \eqref{eq_NE_constraint} guarantees that the speaker responds optimally to the listener's strategy. \eqref{eqn_NE:2} states that the strategy space $\mathcal{M}_{l,t}$ consists of states $\bmzh_t\!\in\! \mathcal{W}$ that maximize the expected semantic information at the listener and responds optimally to the speaker strategy. Next, we formally define the equilibrium points. 
\begin{definition}\vspace{-1mm}
\label{def_localNashEq}
    As defined in \cite{RatliffTAC2016}, we define a strategy $(\pi_{s,t}^*,\pi_{l,t}^*)$ as a \emph{local NE} for \eqref{eq_NashEq_formulation} if there exist open sets $\mathcal{M}_{g_r} \subset \mathcal{M}_{r,t}$ such that $\pi_{r,t}^* \in \mathcal{M}_{g_r}$, where $r \in \left\{s,l \right\}$, and satisfies
\begin{subequations}
  \vspace{-2mm}  \begin{align}
    &-\mathbb{E}_{\bmu_t}\left[S_s(\bmz_t;\bmu_t\mid {\pi}_{s,t}^*,[\bmz_{t-1}],c)\right]  \\ &\geq -\mathbb{E}_{\bmu_t}\left[S_s(\bmz_t;\bmu_t\mid {\pi}_{s,t},[\bmz_{t-1}],c)\right],\,\, \forall\, \pi_{s,t}\in \,\mathcal{M}_{g_s}(\pi_{l,t}^*),\label{eqn:1}\\
        &\mathbb{E}_{\bmu_t}\left[S_l(\bmzh_t;\bmu_t\mid {\pi}_{l,t}^*,\pi_{s,t}^*,[\bmzh_{t-1}],c)\right]\\&\geq \mathbb{E}_{\bmu_t}\left[S_l(\bmzh_t;\bmu_t\mid {\pi}_{l,t},\pi_{s,t}^*,[\bmzh_{t-1}],c)\right], \,\,\forall\, \pi_{l,t}\in \,\mathcal{M}_{g_l}(\pi_{s,t}^*).\label{eqn:2}
    \vspace{-2mm}\end{align}
\vspace{-2mm}\end{subequations}
When $\mathcal{M}_{g_r} = \mathcal{M}_{r}, \forall r$, then the equilibrium is unique.
\vspace{-0mm}\end{definition}
Next, we evaluate the equilibrium points corresponding to \eqref{eq_NashEq_formulation}. 
The speaker in this communication instance aims to minimize the average amount of transmitted semantic information carried by $\bmu_t$, represented by $S_s(\bmz_t;\bmu_t\mid {\pi}_{s,t},[\bmz_{t-1}],c)$. However, the speaker cannot independently optimize this since its strategy space $\mathcal{M}_{s,t}$ depends on the listener inference distribution via the constraint \eqref{eq_NE_constraint}. The mutual understanding evolves via the two-player game proposed in \eqref{eq_NashEq_formulation}, where the listener communicates about the logical formulas it wants to evaluate, and the speaker responds by providing an encoded state description $\bmu_t$ corresponding to the question. 
We assume that $\bmq_t$ and $\bmu_t$ are recovered without any error at the speaker and listener, respectively (reliable wireless links during the emergent language phase). 

\vspace{-3mm}
\subsection{Characterization of the Generalized Local NE}
\label{eq_localNE_char}
\vspace{-1mm}

To find the generalized local NE, one conventional approach is AM between speaker and listener objectives. AM is an iterative procedure for solving the NE objectives jointly over all variables by alternating maximizations over the individual subsets of variables (two NE distributions and Lagrange multipliers). The estimated NE distributions using AM can be computed as follows.
\begin{proposition}
\vspace{-1mm}\label{Solution_NashEq}
    The NE strategies in  \eqref{eq_NashEq_formulation} can be obtained as the following alternating updates. The speaker's strategy is obtained as \eqref{eq_speakTransmit_sol}
  \begin{figure*} \vspace{-1mm} \beq
 \vspace{-1mm} \begin{array}{l}
\pi_{s,t}^* = \pi(\bmu_{t-1})\exp\left(\frac{-1}{S_s(\bmz_t\mid [\bmz_{t-1}],c)\pi(\bmz_t\mid c)}(\lambda_s (c(\bmu_t) - \log \pi_{l,t})-\alpha_{\bmz_t})-1\right). 
\end{array}
\label{eq_speakTransmit_sol}
 \vspace{-1mm} \eeq\end{figure*}
and the listener's inference distribution is \eqref{eq_listInf}.
\begin{figure*}\vspace{-1mm}\beq
\begin{array}{l}
\pi_{l,t}^* = \frac{1}{\pi(\bmu_t)}\exp\left(\frac{-1}{\pi\left(\bmu_t\mid [\bmz_t]\right)S_s({\bmzh}_t\mid [\bmzh_{t-1}],c)Z_{\widehat{z}_tz_t}} \left(\alpha_{\bmu_t}-\pi(\bmu_t\mid [\bmz_t])S_s({\bmzh}_t\mid [\bmzh_{t-1}],c)Z_{\widehat{z}_tz_t}\log\pi_{l,t-1}\right)\right).
\end{array}
\label{eq_listInf}
\eeq
\end{figure*}
\end{proposition}
\begin{IEEEproof}
 \vspace{-0mm} The proof is provided in Appendix~\ref{Proof_Solution_NashEq}.
\end{IEEEproof}

The node policies \eqref{eq_speakTransmit_sol} and \eqref{eq_listInf} suggest that the encoded signal depends on the semantics conveyed by the state description and not on the syntactic representation. Intuitively, the emergent language can represent two state descriptions $\bmz_1,\bmz_2$ leading to the same logical conclusions using the same $\bmu$. Moreover, the speaker transmit policy \eqref{eq_speakTransmit_sol} tends to choose a transmit message $\bmu$ that requires a smaller transmission cost and causes the listener to be less surprised. {If the semantic information in $\bmz$ is high, then the slope parameter $\frac{1}{S_s(\bmz_t\mid [\bmz_{t-1}],\bmc)\pi(\bmz_t\mid \bmc)}$ becomes low. This also suggests that $\pi_{s,t}$ will be very low for short-length codewords, due to higher listener ambiguity. Hence more bits are required to represent $\bmz_t$ with high semantic information.} When $\bmz_t$ provides no new semantic information, the speaker policy's slope becomes infinite due to considering past state descriptions $[\bmz_{t-1}]$ in player strategies. This causes $\pi_{s,t}$ to converge to a Dirac delta distribution, resulting in zero-cost transmission of $\bmu_t$, and no signal is sent. As a result, there is reduced transmission compared to a classical system that doesn't account for semantic history. \eqref{eq_listInf} can be intuitively interpreted as choosing the $\pi_{l,t}$ with its peak close to $\bmzh_t$ that maximizes the average semantic information extracted by the listener for any $\bmu_t$. 
Algorithm~\ref{alg_ta_CSG} summarizes our proposed AM solution. Next, we prove local convergence of the resulting AM updates. 
\vspace{-0mm}\begin{proposition}
\vspace{-0mm}\label{Solution_NashEq_Convergence}
The AM updates obtained in \eqref{eq_speakTransmit_sol} and \eqref{eq_listInf} for the two-player signaling game proposed converges to a local Nash equilibrium as defined in Definition~\ref{def_localNashEq}. 
\vspace{-0mm}\end{proposition}
\vspace{-0mm}\begin{IEEEproof}
    The objective functions in \eqref{eq_NashEq_formulation} are jointly non-concave w.r.t the combined strategy $(\pi_{s,t},\pi_{l,t})$. Hence, the resulting NE cannot be unique. However, for a fixed listener policy, the speaker objective function \eqref{eqn_NE:1} is concave and hence leads to the global optimum, thus satisfying \eqref{eqn:1} for the strategy $\pi_{s,t}^*$. The listener objective function in \eqref{eqn_NE:2} is non-concave w.r.t $\pi_{l,t}$ and can be written as summation of a concave and convex part. To solve this, we computed an approximate function $\widehat{F}_l(\pi_{l,t},\pi_{s,t})$ (defined in \eqref{eq_pi_l_t_approx}) to the original listener objective function \eqref{eqn_NE:2} by approximating the convex part using the first order Taylor series expansion. It can be shown that the approximate function \eqref{eq_pi_l_t_approx} satisfies 
   $ {F}_l(\pi_{l,t},\pi_{s,t}) \geq \widehat{F}_l(\pi_{l,t},\pi_{s,t})$.
    The resulting lower bound is concave \cite{HiriartSpringer1985}; hence, the achieved solution \eqref{eq_listInf} is a local optimum of the original listener objective function. Hence, an AM update between the speaker and listener policy here leads to local NE. The proof for convergence of the AM updates is classical \cite{HongLuoSPM2015}.
\vspace{-0mm}\end{IEEEproof}
\setlength{\textfloatsep}{0pt}
\begin{algorithm}[t]\scriptsize
\caption{Proposed Two-Player Contextual Signaling Game}\label{alg_ta_CSG}
 \textbf{Given:} Sample $\tau_t \sim p(\mathcal{\tau}),p(\mathcal{C}), c(\bmu_t)$:  the  task and distribution over contexts (subscript $t$ represents start time of a task), ${\pi_{s,t-1}^*,\pi_{l,t-1}^*}$\vspace{-0mm}\\ \vspace{-0mm}
\textbf{Initialize:} \mbox{Choose larger values for } $\lambda_s, \balpha_s,\balpha_l$. 
\begin{algorithmic} 
\vspace{-0mm}\STATE \hspace{0.05cm} \textbf{if} ($t > 0$)
\vspace{-0mm}\STATE \hspace{0.5cm} Initialize $\pi_{s,t}=\pi_{s,t-1}^*,\pi_{l,t}=\pi_{l,t-1}^*$ {(starts from the converged values \\\hspace{0.5cm} in previous training instance)}
\vspace{-0mm}\STATE \hspace{0.05cm} \textbf{else}
\vspace{-0mm}\STATE \hspace{0.5cm} Choose uniform distributions for $\pi_{s,t},\pi_{l,t}$.
\vspace{-0mm}\STATE \hspace{0.05cm}  \textbf{end if}
\vspace{-0mm}\STATE \hspace{0.05cm}\textbf{ while} AM not converged \textbf{do}
\vspace{-0mm}\STATE \hspace{0.5cm}  Sample the context $c \sim p(\mathcal{C})$
\vspace{-0mm}\STATE \hspace{0.5cm} \textbf{ for all } $c$ \textbf{do}
\vspace{-0mm}\STATE \hspace{1.2cm} {Evaluate $\pi_{s,t},\pi_{l,t}$ using \eqref{eq_speakTransmit_sol} and \eqref{eq_listInf}, respectively}
\STATE \hspace{1.2cm} Update $\lambda_s, \balpha_s,\balpha_l$ using bisection method  \cite{BoydCUP2004} such that the respective \\ \hspace{1.2cm} constraints
 are met.
\vspace{-0mm}\STATE \hspace{0.5cm}  \textbf{end for}
\vspace{-0mm}\STATE \hspace{0.08cm}\textbf{ end while}
\vspace{-0mm}\STATE \hspace{0.08cm}\textbf{Output} $\pi_{s,t}^*,\pi_{l,t}^*$.
\end{algorithmic}
\label{algo1}  
\vspace{-1mm}\end{algorithm}
\vspace{-0mm}

\vspace{-0mm}
\subsection{Analysis of the Signaling Game Equilibria for Emergent Language}
\vspace{-0mm}

We have so far looked at computing the encoder and decoder probability distributions using the signaling game in Section~\ref{eq_localNE_char}. Next, we look at how the speaker chooses the transmit signal and the listener maps the received signal to the state description given these distributions. Naive approach would be to consider the signals ($\bmu,\bmzh$) as those which maximize the respective probabilities. However, as we show here, that need not be the optimal solution (in terms of the speaker transmitting less and the listener extracting maximum). For notational simplicity, we remove the subscript $t$.
\vspace{-0mm}\begin{definition}\vspace{-0mm}
An equilibrium $(\pi_{s}^*,\pi_{l}^*)$ is a \emph{separating equilibrium} \cite{RatliffTAC2016} if each state
$\bmz$ is encoded as different $\bmu$. That is, $\mathcal{W}$ can be partitioned into regions $\mathcal{W}_k$ such that $\!\sum\limits_{\bmz\in \mathcal{W}_k}\!\pi(\bmu_k\mid \bmz)\!=\!1$. A \emph{pooling equilibrium} occurs if the same $\bmu$ is sent for all  $\bmz$. When $\abs{\mathcal{U}} \!<\! \abs{\mathcal{W}}$, with the same signal transmitted for multiple states but not for all, the resulting equilibrium is a \emph{partial pooling equilibrium}.
\vspace{-0mm}\end{definition}
Using these definitions, we analyze the speaker and listener strategies for different separating and pooling signaling equilibria, as a function of system parameters $\lambda_s$  and $\abs{\mathcal{U}}$.
\begin{theorem}\vspace{-0mm}
\label{analysis_NashEq}
\vspace{-0mm}When the language emerges at the convergence of algorithm~\ref{alg_ta_CSG}, the following equilibria occurs.
\begin{enumerate}\vspace{-1mm}
    \item For $\lambda_{s} \in [0,1]$, the minimum semantic information at speaker side can be achieved by choosing uniformly random the transmit message. Hence, the optimal policies are $\pi(\bmu\mid \bmz) = \frac{1}{K}, \pi(\bmz\mid \bmu) = \pi(\bmz)$ and $H_s(\bmu) = \log K$, where $K$ denotes the cardinality of $\mathcal{U}$. $H(\bmx)$ is the Shannon entropy here.\vspace{-0mm}
    \item For $\lambda_s \!>\! 1$,  the speaker objective can be maximized by choosing a bijective transformation from $\bmz$ to $\bmu$ for the transmit policy (means $\abs{\mathcal{U}} \!=\!\abs{\mathcal{W}} $). This is the perfect signaling or separating equilibrium.  
    \vspace{-0mm}\item For the case of pooling signaling equilibrium, the speaker may send the same $\bmu$ for all $\bmz$ ($\abs{\mathcal{U}}=1$). While this scheme may achieve the minimum semantic information at the speaker, it does not provide any semantic information at the listener and hence it is not suitable for ESC.\vspace{-0mm}
    \item In partial pooling, the listener extracts maximum semantic information when the speaker partition its semantic category space into a Voronoi tessellation, where each $\bmu$ corresponds to a distinct partition.  
\vspace{-1mm}\end{enumerate}
\vspace{-1mm}
\end{theorem}
\vspace{-1mm}\begin{IEEEproof}
   \vspace{-0mm} Proof is provided in Appendix~\ref{proof_analysis_NashEq}.
\end{IEEEproof}
From Theorem~\ref{analysis_NashEq}, for an ESC system where $\abs{\mathcal{U}} < \abs{\mathcal{W}}$, the partial pooling equilibrium is more realistic. An optimal partition at the speaker is such that $\bmu_k$ is transmitted whenever the average semantic information extracted at the listener is the highest among all possible partitions $\mathcal{W}_k$. {Analytically, this can be represented as choosing $\bmu_t = \argmax\limits_{\bmu_t^{\prime}} S_l(\widehat{\bmz}_t;\bmu_t^{\prime}\mid \bmz_t,[\bmzh_{t-1}],c,\pi_{l,t})$.} The proposed ESC system can compute $\pi_{l,t}^*$ using the listener's surprise ($-\log\pi_{l,t}^*$) fed back during the emergent language. Hence, the average semantic information extracted by the listener for different possible partitions can be evaluated. This makes it possible for the speaker to choose the Voronoi tessellation for the encoded signal. In this case, given a Voronoi tessellation of $\mathcal{W}$, the {listener} in turn can choose an
$\widehat{\bmz} \in \mathcal{W}$ for each signal $\bmu_k$ that describes the average state in $\mathcal{W}_k$ optimally (in semantic sense and given the environment or prior for $\bmz$). An optimal
interpretation consists thus of Bayesian estimators for each cell $\mathcal{W}_k$, that is, $
\arg \min_{z\in \widehat{\vspace{-4mm}\mathcal{W}}}\int_{\bmWh} \norm{\bmz-\bmzh}_S \pi(d\bmzh\mid \bmu), $
where $\widehat{\mathcal{W}}$ represents the space over which $\pi(d\bmz\mid \bmu)$ is defined.

\vspace{-4mm}
\section{ NeSy Reasoning for Optimized Semantic Reliability}
\label{GFlowNet}
\vspace{-2mm}

We next 
look at the reasoning neural network (NN) components at the speaker and listener. In short, during the data transmission phase, the speaker's reasoning component allows the computation of the optimal set of entities and relations that causally explain the data. The awareness of how causal description gets used by the listener enables the speaker to encode with minimal bits during the emergent language phase discussed in Section~\ref{EmergentLanguage}. 
Our next key question is this: What is an optimal reasoning scheme at both the nodes such that it enables the listener to reliably reconstruct the semantic aspect (logical conclusions) considering that the transmitted signal is distorted via the underlying channel $p(\bmx\mid\bmu)$ ? Before discussing the proposed reasoning components, we first assess how reasoning helps our ESC system reduce the average number of bits for transmission and achieve better semantic reliability than classical systems. \emph{Semantic reliability} here is defined as the accuracy in terms of the semantics reconstructed at the listener compared to the intended semantics transmitted from the speaker.

\vspace{-1mm}
\subsection{Average Semantic Representation Length for Classical and Emergent Language based ESC}
\vspace{-1mm}

Let $c$ be the context, drawn from $p_c(c)$. Depending on $c$, it is possible that a certain entity relation pair (or entity relation sequence of pairs) is not valid. Hence, the possible state sequences that generate the data will be a subset of $\mW$. Depending on this subset, the transmitted vocabulary $\mathcal{U}$ size can be dynamically adjusted if the emergent language is aware of the context $c$. Motivated by the importance of communication context significance in the optimal vocabulary design, we derive the following result. 
\vspace{-1mm}\begin{theorem}\vspace{-1mm}
\label{theorem_se_representation}
For a particular syntactic space, $\mathcal{W}$ and context distribution $p(c)$ over $\mathcal{C}$, the average number of bits to represent the state description in an ESC system can be bounded as \eqref{eq_RepLength_Theorem}.
\begin{figure*}\vspace{-1mm}\beq\vspace{-1mm}
\sum\limits_{c\in \mathcal{C}}\pi( c) H(\bmz_i\mid c) \leq \sum\limits_{\bmu_i \in \mathcal{U}}\pi(\bmu_i)l_i   \leq -\sum\limits_{c}\pi( c) \sum\limits_{\bmz_i}\pi(\bmz_i\mid c)\lceil\log \pi(\bmz_i\mid c)\rceil
,\label{eq_RepLength_Theorem}
 \vspace{-0mm} \eeq
 \end{figure*}
where $l_i$ represents the number of bits to represent $\bmu_i$. For a classical system, the bounds are \eqref{eq_classicalbound}.
\begin{figure*}\beq
H_c(\bmz_i) \leq \sum\limits_{\bmu_i \in \mathcal{U}}\pi(\bmu_i)l_i  \leq  \max\limits_{c}\sum\limits_{\bmz_i}\pi(\bmz_i\mid c)\left[\sum\limits_{\bms_i \in \bmz_i}\lceil\log \pi(\bms_i\mid c)\rceil\right].
 \vspace{-1mm}\label{eq_classicalbound}\eeq\end{figure*}
 \vspace{-0mm} \end{theorem}
  \vspace{-0mm}\begin{IEEEproof}
    \vspace{-0mm} See Appendix~\ref{appendix_proof_theorem_se_rep}.
   \vspace{-0mm}\end{IEEEproof}
   {From Theorem~\ref{theorem_se_representation}, as explained in Appendix~\ref{appendix_proof_theorem_se_rep}, it can be shown that for an ESC system, the lower and upper bounds for a physical representation of the semantics are smaller to those in a classical system (transmitting a compressed version of the entities), justifying its transmission efficiency.} Next, we look at whether the listener's semantic reliability improves by incorporating reasoning at the nodes.
\vspace{-1mm}\begin{theorem}
     \vspace{-1mm} \label{theorem_se_error}
For a given representation space $\mathcal{U}$, the lower bound on the semantic error probability ($S_e-$representing reliability) is always less than or equal to the lower bound on the probability of bit error ($P_e$) measure achieved using classical communication system.
\beq
\vspace{-1mm}
\begin{array}{l}
P_e \geq \frac{H(\bmzh_c\mid \bmz_c) - 1}{\log \abs{\mathcal{W}}} , \,\,\,  S_e \geq \frac{H(\bmz\mid \bmzh) - H(\bme\mid \bmzh)}{\log \abs{\mathcal{W}}}  \\
\mbox{where}, \,\frac{H(\bmzh_c\mid \bmz_c) - 1}{\log \abs{\mathcal{W}}} \geq \frac{H(\bmz\mid \bmzh) - H(\bme\mid \bmzh)}{\log \abs{\mathcal{W}}}. 
\vspace{-2mm}\end{array}
 \vspace{-1mm}\eeq
 \vspace{-1mm} \end{theorem}
\begin{IEEEproof}\vspace{-0mm}
    See Appendix~\ref{proof_theorem_se_error}. Here, $\bme$ represents the error in semantics recovered at the listener, and $\bmz_c, \bmzh_c$ means the state descriptions for the classical system. 
\end{IEEEproof}
Theorem~\ref{theorem_se_error} shows that inducing reasoning at speaker and listener can improve semantic reliability compared to a classical system that uses the same number of bits to communicate. In other words, the ESC system could choose to transmit at a higher rate than a classical system to achieve similar semantic reliability. Next, we look at the structure of reasoning components at the speaker and listener that can help the ESC system achieve gains discussed above.

\vspace{-3mm}
\subsection{Semantic State Descriptor at Speaker}
\vspace{-1mm}

The causal state description at the speaker is represented as $\bmz= \left[\bms_0,\bms_1\cdots ,\bms_f\right]^T\!$ and learned as the posterior distribution \eqref{eq_DAG_causality}. 
The GFlowNet architecture that we use to learn this posterior is based on \cite[Fig. 2]{DeleuArxiv2022}. Each directed edge (between $\bms_i$'s) is embedded using its source and target's embeddings (\emph{grounded} value), with an additional vector indicating whether the edge is present in $G$. These embeddings are fed into a linear
transformer, with two separate output heads. The first head gives the probability of adding a new edge $P_{\btheta}(G^{\prime}
\!\mid \!G, \neg \bms_f )$, using the mask $M$ associated with $G$ to filter out invalid actions. $G^{\prime}$ is the newly formed DAG. The second head gives the probability of terminating the trajectory, $P(\bms\!=\!\bms_f\!\mid\! G)$. Once the terminal state is created, we would have formed a probabilistic graphical model to represent the data's causal structure ($\bmz$) using the explainable AI tools of GFlowNet.   

\vspace{-3mm}
\subsection{Listener Semantic Reasoning }
\vspace{-1mm}

\begin{figure}[t]\vspace{-4mm}
\centerline{\includegraphics[width=3.5in,height=2.2in]{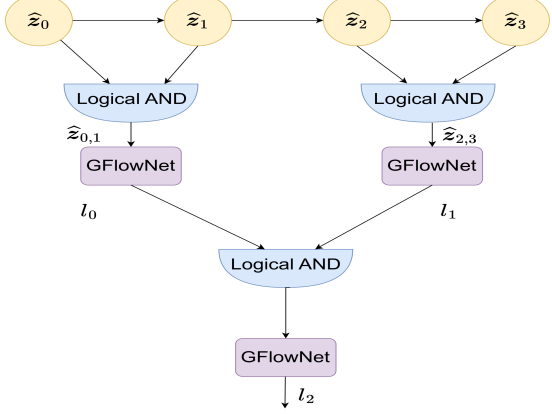}} \vspace{-2mm}
\caption{\scriptsize GFlowNet + LNN for logical theorem proving, a symbolic process.}
\label{Fig_Theorem_Proving1}
\vspace{-0mm}\end{figure}

At the listener, 
the received signals pass through the decoder which maps the received signal to representation space $\mathcal{U}$ using a single layer deep NN (DNN) with parameters $\bmeta$ and $\widehat{\bmu} \!= \!g_{\bmeta}(\bmx)$, with $\widehat{\bmu} \in \mathcal{U}$. 
From the decoded signals $\bmuh$, the semantic state and context are extracted using the inference distribution $\pi_l$. From the extracted semantic state $\bmzh$, the listener intends to evaluate the logical formulas $\bphi_i$. In other words, the listener would want to infer the possible world states arising from the observed event at the speaker. 
We now explain how we can exploit the \emph{chain of thought reasoning} \cite{WeiChainReasonArxiv2022} to evaluate a logical formula given a complex state description that is unseen during training. 
{\subsubsection{Evaluating the logical formulas}
We now explain how the listener uses semantic reasoning so as to helps generalize to unseen scenarios aided by the compositional emergent language. Utilizing the extracted state descriptions, we propose to use a combination of GFlowNet and NN based logical connectives (inspired from LNN) to evaluate the logical formulas. Fig.~\ref{Fig_Theorem_Proving1} serves as an example to illustrate this procedure, and it is noteworthy that the architecture can be generalized for values of $R$ greater than $4$ by concatenating similar connectives and GFlowNets, using $\bphi_i$ defined in \eqref{eq_phi_i}.
Assume that the state description $\bmzh$  can be split into $R$ simpler state descriptions, i.e., $\bmzh = (\bmzh_1,\bmzh_2,\cdots,\bmzh_R)$. We combine the consecutive state descriptions $\bmzh_k$ and $\bmzh_{k+1}$, to obtain a possible outcome, $l_j(\bmzh_{k,k+1})$, which gets evaluated by GFlowNet. The combined state $\bmzh_{i,i+1}$ is computed as
\vspace{-1mm}\beq
\vspace{-1mm}\bmzh_{i,i+1} = f(\beta-\bmw_i^T(1-\bmzh_i)-\bmw_{i+1}^T(1-\bmzh_{i+1})),
\label{eq_z_i_1}
\vspace{-1mm}\eeq
where $f(z)$ is chosen as the logistic activation function \cite{RiegelLNN2020} and the coefficients $\beta$ and $\bmw_i$ are learned during training.  
Further, using the chain-like rules in Definition~\ref{def_chain_like_rule}, we can write the complex logical formula $
\bphi_i = (l_1^{(i)},l_2^{(i)},\cdots, l_{R-1}^{(i)}),$ where $R=4$ in Fig.~\ref{Fig_Theorem_Proving1}. The chain-like rules can be applied since we can use the property of functors in category theory (see Appendix~\ref{app_catTheory}) to state that if two objects $\bmzh_1$ and $\bmzh_2$ in the syntax category are related, then the corresponding outcomes $l_1^{(i)}$ and $l_2^{(i)}$ are also related. }
In short, this functionality highlights GFlowNet's logical theorem proving capability, the symbolic component for the proposed NeSy AI algorithm. 

\vspace{-3mm}
\subsection{Semantic Distortion and Semantic Similarity}
\label{SD}
\vspace{-1mm}

The channel imperfections result in an error in the reconstructed $\bmz_t$; leading to erroneous logical conclusions by the listener, and this error in $\bphi$, captured below, is called \emph{semantic distortion}. 
\vspace{-3mm}\beq
\vspace{-2mm}
E_t(\bmz_t,{\widehat{\bmz}}_t) = \dsum\limits_{i=1}^L\norm{\bphi_{i}(\bmz_t)-{{\bphi_i}(\bmzh_t)}}^2.
\vspace{-0mm}
\eeq
We also define the error measure in terms of the difference in semantic information conveyed by the speaker and that learned by the listener,  $E_t({S}_t,{\widehat{S}}_t) = \abs{S_t-\widehat{S}_t}^2.$
These semantic distortion measures quantify the amount of semantic information that the listener can extract from a corrupted message.

Next, we define the concept of \emph{semantic similarity} which significantly differs from the bit error rate (BER) of classical systems and is required for defining the semantic reliability constraint. In a sense, we may send a message that experiences a high BER, but {the reconstructed $\bmzh_t$} may be sufficient to recover the intended meaning. For example, say the listener receives the English sentence ``Aln the vhicles have sped lss thn 50 km/hr" instead of ``All the vehicles have speed less than 50 km/hr" (actually intended by the speaker). The received text has high BER (called an error at syntactic level \cite{ChaccourArxiv2022}), i.e., typos, but its semantic content is still rich and can be recovered. Several sentences can correspond to the same semantic space in our ESC model. The decoded state description at the listener restricted to this semantic message space (say represented as $\mathcal{S}_t$ corresponding to message $\bmz_t$) can represent the same semantic content and they are semantically similar. Hence, we can achieve the same reliability. This semantic message space (representing semantic similarity) corresponding to $\widehat{\bmz}_t$ can be described as $
E_t(\bmz_t,{\widehat{\bmz}}_t)  \leq \delta,\,\, \mbox{s.t.}\,\, E_t({S}_t,{\widehat{S}}_t) = 0.
$
From a category theory perspective, we can interpret the semantic message space as the set of all $\bmz_t$ such that the representable copresheaves are identical. Mathematically, we can write $\mathcal{S}_t = \left\{\forall \,\widehat{\bmz}_t \,\,\,s.t.\,\,\, H^{\widehat{\bmz}_t} = \mH^{\bmz_t} \right\}.$
Moreover, note that $\bphi(\bmzh) \in H^{\widehat{\bmz}_t}$.
Compared to the classical BER measure, the semantic distortion $E_t(\bmz_t,{\widehat{\bmz}}_t)$ need not be restricted by an arbitrarily low value to achieve a reliability close to $1$. 

\vspace{-3mm}
\subsection{Causal Influence of Communication Channel on ESC and Semantic Effectiveness}
\vspace{-1mm}

Further, we introduce metric $C_t$, interpreted as the causal influence on ESC, and it captures the causal impact of the speaker's message (via listener's actions) as observed through a channel with a response characterized using $p(\bmx_t\mid \bmu_t)$. In short, $C_t$ captures the semantic effectiveness (inversely proportional to $C_t$) of the transmitted message to the end user. We define 
\vspace{-1mm}
\beq
\vspace{-1mm}
\begin{array}{l} \small
C_t(\bmz_t,\bmz_t(\bmeta)) \\ =  D_{KL} \left(p(\bma_t \mid \bmz_t) \,||\, \sum\limits_{\bmx_t} p(\widehat{\bmz}_t \mid \widehat{\bmu}_t )p({\bmx}_t \mid \bmu_t )p(\bma_t\mid\widehat{\bmz}_t)\right) ,
\end{array}
\vspace{-0.5mm}
\eeq
where $D_{KL} (p||q)$ represents the KLD between $p$ and $q$. {We define semantic effectiveness as $C^{\prime}_{t} = \frac{1}{1+C_t}$, where $C_t$ is a measure of the discrepancy between the ideal action probability and the probability obtained from the reconstruction of the received semantics. A value of zero for $C_t$ indicates that the $C^{\prime}_{t}$ is highest, corresponding to the ideal case where the reconstructed semantics match the ideal action probability. As the value of $C_t$ increases, the {listener's} actions become increasingly distant from the ideal case, resulting in lower semantic effectiveness.} Since the above semantic effectiveness metric captures the channel effect on the transmitted semantics, the learned GFlowNet (for listener side) parameters is robust to the channel errors (in terms of the best the listener can do).

\vspace{-3mm}
\subsection{Proposed ESC Reasoning Network Formulation and Solution}
\vspace{-1mm}

By leveraging the semantic metrics defined above, we can now rigorously formulate our objective. The speaker aims to choose an optimal state description to represent the causal reasoning at the transmit side and the listener wants to increase semantic effectiveness at its side to achieve a desired semantic reliability. Using the measures discussed in Section~\ref{SD}, \emph{semantic reliability} (i.e.,$=p\left(E_t(\bmz_t,{\widehat{\bmz}}_t(\bmeta) ) < \delta \right)$) is defined so that the listener can reliably reconstruct all the logical formulas contained in the decoded state description. Compared to classical reliability measures, in semantics, we can recover the actual meaning of the transmitted messages even with a larger BER, as long as the semantic distortion is within a limit. We formally pose our problem for speaker ($ \mathcal{P}_S$) and listener ($ \mathcal{P}_L$) causal reasoning as {
\begin{subequations}\vspace{-1mm}
\begin{align}
&\mathcal{P}_S:\left[G^*,\btheta^*\right]  = \hspace{1mm} \argmin \limits_{\btheta,G}
 L_{\btheta}\left(\tau_s\right)   \\ \vspace{-1mm}
 &\mathcal{P}_L:   \left[\widehat{G}^*,\bpsi^*,\bmeta^*,\beta^*,\bmw_i^*\right] \\ & \hspace{19mm}=  \argmin \limits_{\bpsi, \bmeta,\widehat{G},\beta,\bmw_i \forall i}
C_t\left(\bmz_t,\bmzh_t(\bmeta)\right) + L_{\bpsi}\left(\tau_l\right)  \\ \vspace{-5mm}
  &\hspace{20mm}\mbox{s.t}\,\,\, p\left(E_t\left({\bmz}_t,{\widehat{\bmz}}_t\left(\bmeta\right) \right) < \delta \right) \geq 1-\epsilon,
\vspace{-1mm}\end{align}
\label{GFlowNetLossFunc}
\vspace{-0mm}
\end{subequations}\!\!\!where $\delta, \epsilon$ can be arbitrarily small values greater than 0. In \eqref{GFlowNetLossFunc}, $\widehat{G}$ denotes the extracted causal graph at the listener side. From the learned NN parameters $\bpsi^*$, the listener reconstructs the state description and evaluates the logical formulas in \eqref{eq_phi_i} using $\beta^*,\bmw_i^*$.} The value
of $\delta$ here depends on the $\mathcal{W}$ space, and hence the NN parameters learned for all task variations with the
same syntactic space remain intact. The loss function to optimize speaker and listener GFlowNet parameters is expressed as
\vspace{-2mm}\beq \vspace{-1mm}
\small
\begin{array}{l}
\!L_{\btheta}(\tau) = \!\!\! \sum\limits_{s^{\prime} \in \tau \neq s_0}\!\!\left(\!\sum\limits_{s,a:\Pi(s,a)=s^{\prime}}\!\!\!\!\!\!\!\!F_{\btheta}(s,a)-R(s^{\prime})-\!\!\!\sum\limits_{a^{\prime}\in \mathcal{A}(s^{\prime})}F_{\btheta}(s^{\prime},a^{\prime})\!\right)^2\!\!.
\end{array}
\eeq
$\Pi(s,a)$ represents the state transition in the DAG.  {For the non-terminal states, we have $R(\bms^{\prime}) = 0$. From an SC system point of view, we choose a reward for terminal states (and hence the state descriptions) such that they provide maximum semantic information given the history of the state descriptions $[\bmz_t]$. Hence, we consider $R(\bms_{N-1}) = S(\bmz\mid c)$, where $\bmz = (\bms_0,\cdots,\bms_{N-1})$.} Flow $F_{\btheta}(s,a)$ is parameterized by GFlowNet's NN and is computed during offline training, and it does not impact the communication overhead. Since the listener is aware of the syntactic space, it can train the speaker's version of GFlowNet offline, without any communication overhead. 

Our objective function \eqref{GFlowNetLossFunc} is a novel rigorous formulation that captures accurately the semantic reasoning capabilities associated with the end nodes. To solve \eqref{GFlowNetLossFunc} whose intent is to achieve high semantic reliability, we use the back propagation to fine tune GFlowNet and DNN weights. Our objective function in \eqref{GFlowNetLossFunc} is convex. This can be easily proven since KLD is convex and the convergence properties of GFlowNet are already proved in \cite{BengioNeurIPS2021}. Hence, the convergence to any global minimum solution is guaranteed for our solution. We train over several mini-batches iteratively and hence the optimization of the network parameters are done over an $\small \mathbb{E}_{e\sim D}[\mathcal{P}_S]$ or $\small \mathbb{E}_{e\sim D}[\mathcal{P}_L]$. Moreover, we consider our GFlowNet to be evolving which means that during the every training instance, terminal state probability $p(s_f\mid G)$ gets updated. The algorithmic details of the proposed NeSy AI based ESC system is described in Algorithm~\ref{alg_1}. Next, we evaluate via simulations the proposed solutions.

{\subsubsection{Algorithm complexity discussion} The GFlowNet architecture used in our proposed ESC is same as in Fig. 2 of \cite{DeleuArxiv2022} that utilizes linear transformers \cite{VaswaniNIPS2017} as its core component. The linear transformer used here only requires $\mO(MN)$ attention layers, where $M$ is the dimension of each entity and $M<N$. This is a significant reduction in complexity compared to other transformer model, whose complexity is $\mO(N^2)$, with competitive performance compared to other transformer models. However, we acknowledge that deploying such models on low-computational-power devices (e.g., sensors) is not feasible. We could possibly perform the training on next-generation smartphones or laptops that may have the enhanced capabilites above or are equipped with powerful GPUs. However, for devices with limited resources, it is recommended to conduct model training on high-performance servers, while the resulting model can be downloaded to edge devices for inference. One advantage of our
overall approach, compared to other state-of-the-art native AI wireless systems, is that capturing
the causal structure leads to greater generalizability. This means that the training effort required
for future tasks is minimal or negligible. As a result, the ultra-high reliability, rate, and ultra-low
latency requirements of future wireless systems are not hindered by the need for retraining. }  
\vspace{-2mm}\setlength{\textfloatsep}{0pt}
\begin{algorithm}[t] \scriptsize{ 
\caption{NeSy Approach towards Semantic Transmission - Training Phase for GFlowNet and LNN}\label{alg_1}
 \textbf{Given:} $\mbox{Channel transition probability} \,\,p(\bmx_t\mid\bmu_t), \bphi_i, R$.\\
\textbf{Initialize:} \mbox{Adjacency matrix } $\bmW^{(0)}$ \,\,\mbox{as all zeros matrix}. Mask $\bmM^{(0)}$ to be constrained to have no loops. \;\textbf{Set:} $\mu = 0.4$.\vspace{-0.5mm} \vspace{-0mm}
\begin{algorithmic} \vspace{-0mm}
\FOR{ minibatch b=1:B}\vspace{-0mm}
\FOR{node $n=$ speaker, listener (Executed parallely)} \vspace{-1mm}
\STATE  Sample mini batch $(b)$ of $N_1$ events from the training set, $e \sim D$. \vspace{-0mm}
\FOR{{ each event in $b$ repeat until terminal state is reached}} \vspace{-0mm}
\IF{$n==$speaker}\vspace{-0mm}
\STATE   Optimize the loss function in $\mP_S$ in \eqref{GFlowNetLossFunc}, using ADAM optimizer \cite{KingmaArxiv2014} with a learning rate of $0.001$. 
\STATE Compute  $P(G^{(t)}\mid G^{(t-1)},\neg s_f)$ and  $P(s_f\mid G^{(t-1)})$, that follows from the GFlowNet architecture in \cite[Fig. 2]{DeleuArxiv2022}.\vspace{-0mm}
\STATE\ELSE\vspace{-0mm}
\STATE   Optimize the loss function in $\mP_L$ in \eqref{GFlowNetLossFunc}, using ADAM optimizer \cite{KingmaArxiv2014} with a learning rate of $0.001$.
\STATE Compute  $P(G^{(t)}\mid G^{(t-1)},\neg s_f)$ and  $P(s_f\mid G^{(t-1)})$ that follows from the GFlowNet architecture in \cite[Fig. 2]{DeleuArxiv2022}.\vspace{-2mm}
\STATE \ENDIF\vspace{-2mm}
\STATE  \IF{$P(s_f\mid G^{(t-1)}) > \mu$} 
\STATE  exit the training for the event \vspace{-1mm}
\STATE  \ELSE\vspace{-0mm}
\STATE Update the node $n$'s GFlowNet parameters, mask and $\bmW^{(t)}$. \vspace{-0mm}
\IF{n==listener}\vspace{-0mm}
\STATE Update the DNN parameters for the listener. 
\STATE Update the LNN parameters and evaluate the logical formulas $\bphi_i$, which is defined in \eqref{eq_phi_i}   and whose architecture follows Fig.~\ref{Fig_Theorem_Proving1}.  \vspace{-3mm}
\STATE \ENDIF
\STATE \ENDIF\vspace{-0mm}
 \ENDFOR \vspace{-0mm}
 \ENDFOR \vspace{-0mm}
 \ENDFOR\vspace{-0mm}\\
 \STATE Use the learned GFlowNet and decoder weights for the data transmission phase.
 \end{algorithmic}\vspace{-0mm}
\label{algo1}  }
\vspace{-0.5mm}\end{algorithm}

\vspace{-2mm}\section{Simulation Results and Analysis}
\vspace{-1mm}

In this section, we first start with a simple illustrative scenario to demonstrate the gains from a causal reasoning based ESC system as derived in Theorems~\ref{theorem_se_representation} and~\ref{theorem_se_error}. {The baselines for comparison are a classical AI based wireless system that directly transmits the features extracted using a convolutional neural
networks (CNN) and to a state-of-the-art SC system that uses reinforcement learning to design the encoder, decoder, and the semantic reasoning paths \cite{XiaoICC2022} (represented as ``Implicit Semantic Communicaton Architecture" in the Legend. The AI architecture utilized in the CNN-based conventional system remains unchanged from \cite{LiangJSTSP2018}, and the entities are encoded directly without any causal extraction on either end.} Further, we present extensive simulation results to assess our ESC system performance.

\vspace{-3mm}\subsection{Illustrative Example for NeSy AI's Potential in Wireless vs Classical AI based Wireless}
\vspace{-1mm}

Our proposed ESC framework can support several wireless applications. Nevertheless, for an illustrative example that is simple to understand, we consider the events as the images observed from a camera at the speaker. Assume that the image's background is irrelevant, and the particular { entity} present (with the possible attributes as in Fig.~\ref{testbed} is of interest to the listener. {We use SHAPES which is a synthetic dataset created by \cite{AndreasCVPR2016}. In this dataset, each image is a $30\times 30$ RGB image showing a $3\times 3$ grid of objects. Each object is characterized by shape, colour and stye as shown in Fig. 5a. This collection comprises complex questions pertaining to simple arrangements of colored shapes. Each question involves two to four characteristics, object types, or relationships. For example: Is there a red shape above a circle? Overall, the dataset contains $244$ questions and $15,616$ images. During training, for every minibatch of the dataset, the listener samples a random question from the dataset that acts as the logical conclusion here.  During the communication phase, the listener employs the decoder distribution $\pi_{l,t}$ to reconstruct the sequence of entities $\bmzh_t$, with each entity corresponding to a specific object. The relations among these entities are then determined using GFlowNet, and logical formulas are evaluated through a combination of GFlowNet and logical neural networks, as discussed in Section IV.C. As an illustrative example, consider that, in task $\tau_1$, learning can occur over contexts described by state descriptions of the form $(a,b,c)$, where the entities $a$, $b$, and $c$ can be one of three distinct values: $\{`shape', `color', `style'\}$. However, `style' is irrelevant to the task, leaving only $16$ relevant state descriptions. Among these state descriptions, from the listener's perspective, suppose there exist four unique conclusions that can be drawn from a subset of four states. Therefore, a speaker aware of the listener's semantics can communicate using only two bits, as states with the same logical conclusions can be encoded as the same transmit signal. In contrast, a naive wireless system unaware of the semantics and redundancy may require up to six bits to communicate all $64$ possible state descriptions.} Considering training complexity, for an ESC system, task $\tau_1$ just requires $8$ communication rounds, wherein each round, the information about a single attribute is communicated from speaker to listener. However, a classical AI system would require to be trained with enough data that involves possible $16$ different objects. As the complexity of the task increases, which implies each context involving $3$ attributes and multiple objects, the training complexity (to achieve required reliability) of the classical AI system increases exponentially. Fig.~\ref{SemReliability_SC_CLassicalAI} compares the semantic reliability for classical and ESC systems with the same amount of training data used, where the ESC is superior due to its generalization capability (with fewer training data). {Additionally, the ESC system proposed in this study exhibits a $63\%$ improvement in semantic reliability compared to implicit semantic communication architecture (at task difficulty $=3$) based on reasoning. This improvement can be attributed to extracting causal reasoning and emergent language components.} 
Fig.~\ref{numBitsSC} conveys the potential gain by using a causal reasoning based ESC system in terms of the average representation length (bits).  {Fig.~\ref{numBitsSC} shows a $23\%$ reduction in semantic representation length for the ESC system compared to the implicit semantic communication architecture.} Here, we used a binary symmetric channel (BSC) to represent the underlying wireless channel for communication. Having understood the fundamental performance gains from a reasoning based system, we further look at evaluating our algorithms.
\begin{figure}[t]\vspace{-1mm}
\centerline{\includegraphics[width=3.4in,height=1.8in]{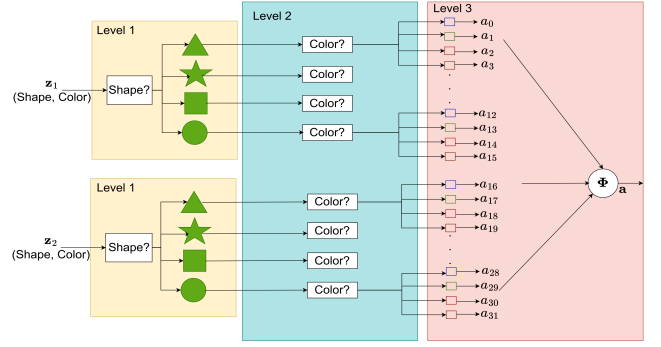}} \vspace{-1mm}
\caption{\scriptsize Example on how logical reasoning (task with two objects present) is performed.}
\label{Fig_Theorem_Proving}
\vspace{-0mm}
\vspace{-1mm}\end{figure}
\begin{figure*}[t]
\vspace{-5mm}
 \begin{subfigure}{.31\textwidth}
{\includegraphics[width=2in,height=1.5in]{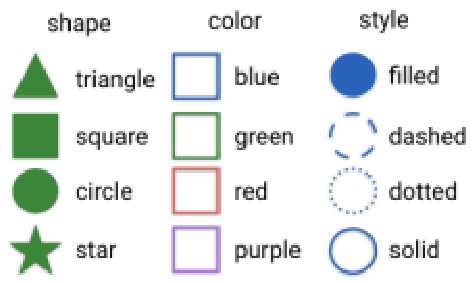}}\vspace{-1mm}
\caption{}
\label{testbed}\vspace{-1mm}
\end{subfigure}
 \begin{subfigure}{.29\textwidth}
\centerline{\includegraphics[width=2.5in,height=1.5in]{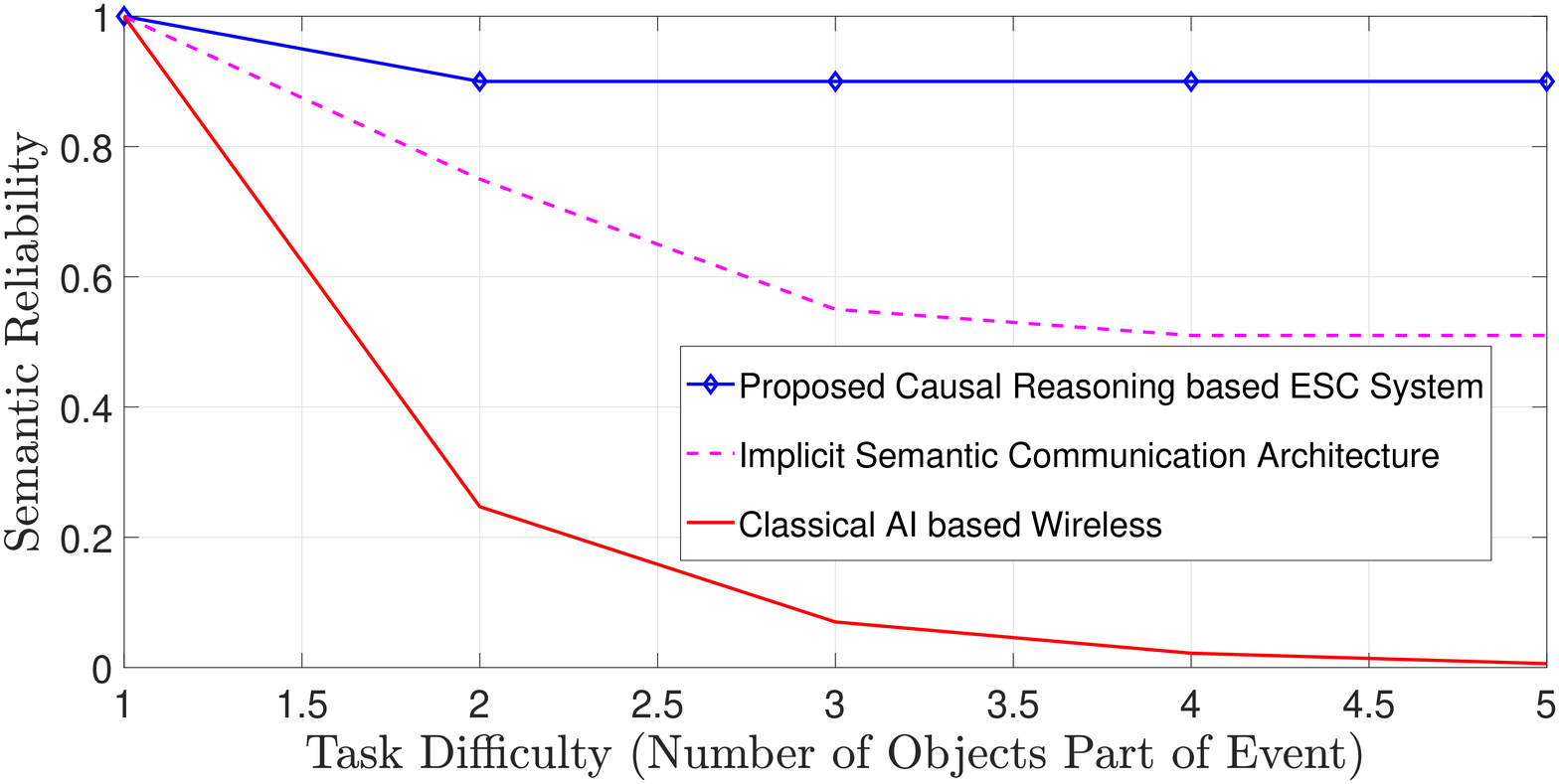}}\vspace{-1mm}
 \caption{}
\label{SemReliability_SC_CLassicalAI}\vspace{-1mm}
\end{subfigure}
\begin{subfigure}{.4\textwidth}
\centerline{\includegraphics[width=2.5in,height=1.5in]{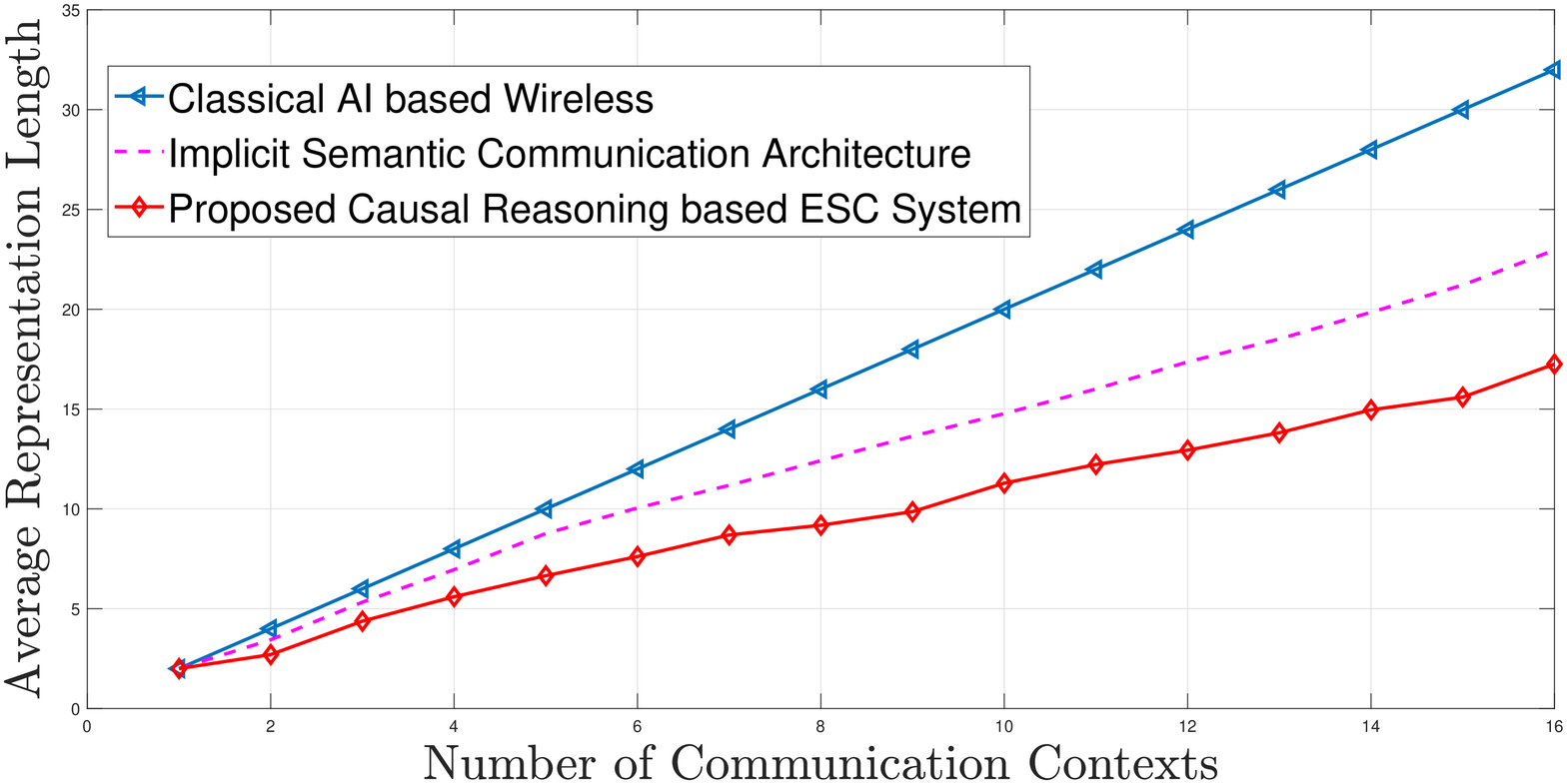}\hspace{-10mm}}\vspace{-1mm}
\caption{}
    \label{numBitsSC}\vspace{-1mm}
\end{subfigure}
\vspace{-2.5mm}\caption{\scriptsize \!(a) The testbed for the illustrative example is the state description grounded in a synthetic
world of objects with $4$ shapes $\!\times\!$ $4$ colors $\!\times\!$ $4$ styles. { (b) Classical AI based system fails with task difficulty (increase in number of objects describing an event). (c) Classical wireless system is designed based on worst case communication context while contextual aware ESC has a dynamically changing representation length, resulting in an efficient transmission.}}
\label{Fig_illustration}
\vspace{-0mm}\end{figure*}

\vspace{-3mm}\subsection{Proposed Contextual Two-Player Signaling Game Evaluation: Convergence and Semantic Reliability}
\vspace{-1mm}

\begin{figure*}[t]
\begin{subfigure}{.31\textwidth}\vspace{-0mm}
{\includegraphics[width=2.1in,height=1.9in]{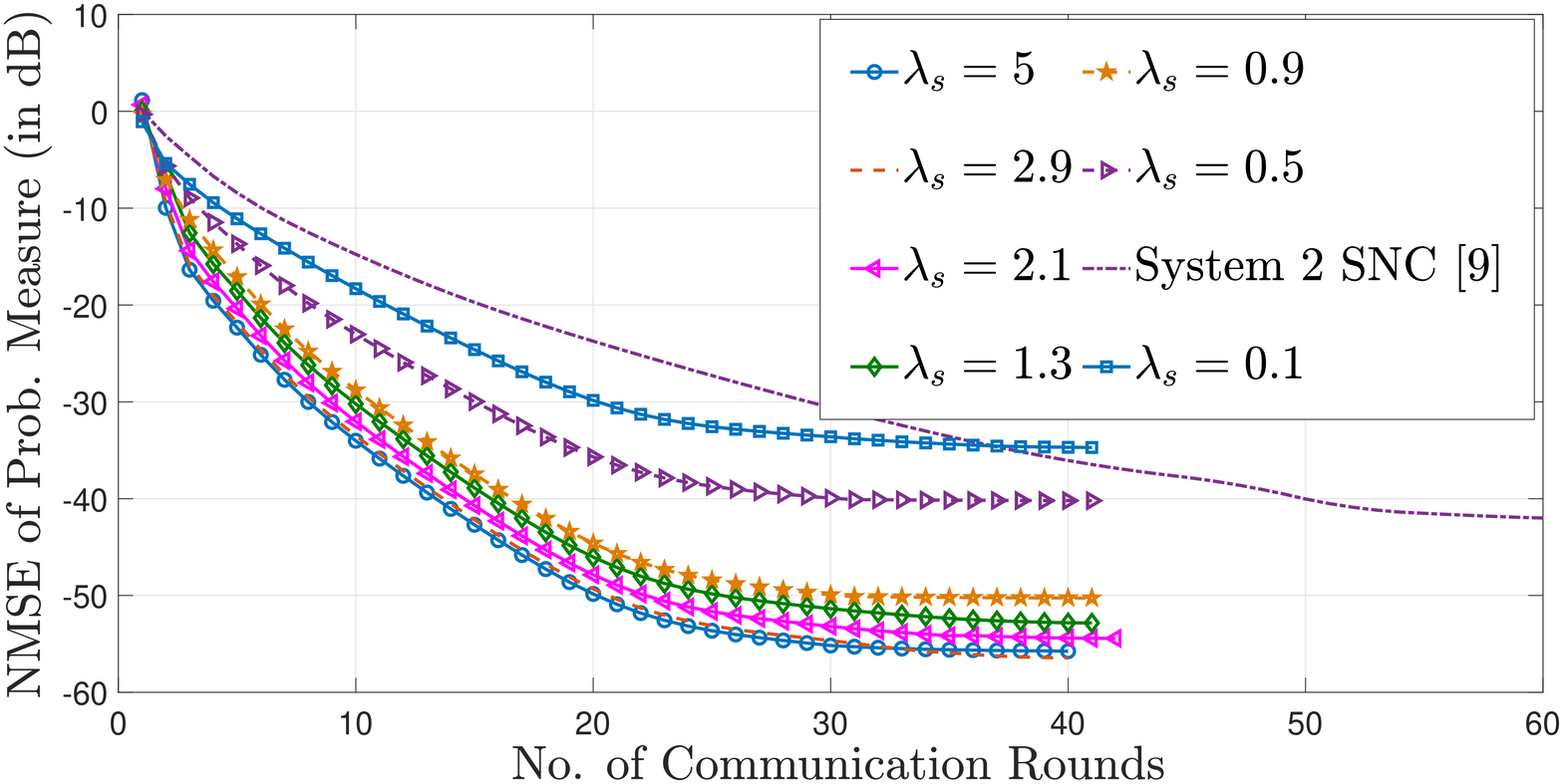}}\vspace{-1mm}
\caption{}
\label{Convergence_algo1}\vspace{-1mm}
\end{subfigure}
\vspace{-0mm}\begin{subfigure}{.30\textwidth}
\vspace{-0mm}{\includegraphics[width=2.2in,height=1.9in]{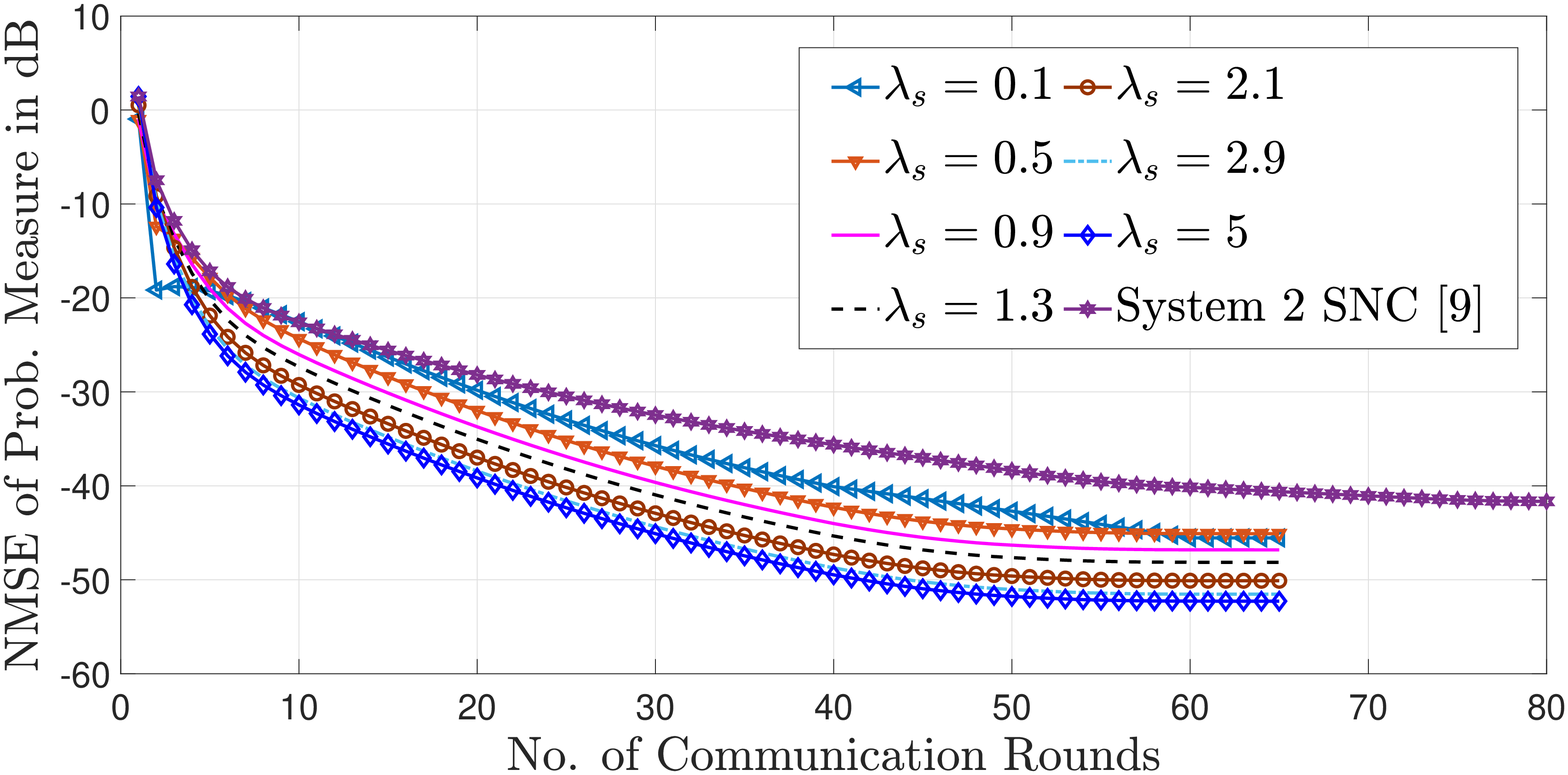}}\vspace{-1mm}
\caption{}
\label{SpeakerPolicyConver_C7}\vspace{-1mm}
\end{subfigure}
\vspace{-1mm}\begin{subfigure}{.39\textwidth}\vspace{-3mm}
\centerline{\hspace{2mm}\includegraphics[width=2.6in,height=1.9in]{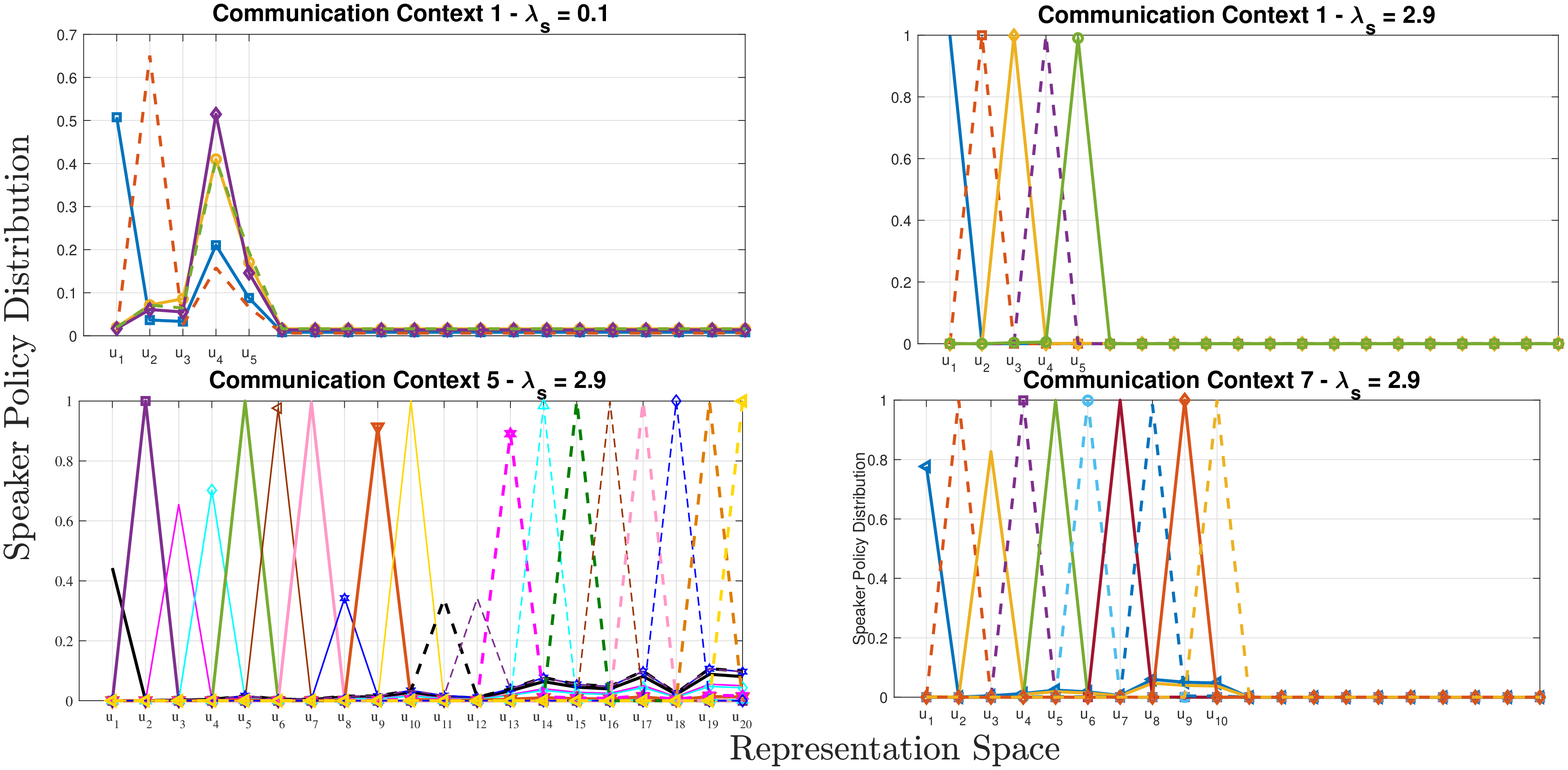}}\vspace{-1mm}
\caption{}
\label{SpeakerPolicy}\vspace{-1mm}
\vspace{-1mm}\end{subfigure}
\vspace{-4mm}\caption{\scriptsize (a) Convergence vs no. of communication rounds, for $\abs{\mathcal{W}} = 100$ and $\abs{\mathcal{U}}=5$. (b) Convergence vs no. of communication rounds, for $\abs{\mathcal{W}}=200, \abs{\mathcal{U}}=20$. (c) Speaker transmit policy for different contexts. Proposed scheme has its $\abs{\mathcal{U}}$ varying w.r.t the communication context (each curve, with distinct color, show $\pi_s$ given a different $\bmz$).}
\vspace{-0mm}
\label{EMLangCOnvergence}
\end{figure*}

To evaluate the convergence of the Algorithm~\ref{alg_ta_CSG}, we consider $\abs{\mathcal{W}} = 100$ and $35$ different communication contexts. Each context sampled uniformly random from the context set can have maximum $20$ states as valid entries and with a minimum of $5$ states per context. The cost of transmission is $c(\bmu_i) = \lceil(i/5)\rceil$. In Fig.~\ref{Convergence_algo1}, we look at the convergence behavior of the proposed algorithm for different $\lambda_s$. To achieve an NMSE of around $-50$ dB, as the figure suggests, we need around $20$ communication rounds during the emergent language phase. Fig.~\ref{SpeakerPolicyConver_C7} shows the varying convergence behavior when the cardinality of $\mathcal{W}$ and $\mathcal{U}$ changes. In Fig.~\ref{SpeakerPolicy}, we present the converged NE speaker policies resulting from our game. Clearly, the resulting speaker policy is context-dependent and hence, the representation space gets optimized dynamically. Also, our scheme converges three times faster than the baseline \cite{Mehdi2021}. 

\begin{figure*}[t]\vspace{-3.5mm}
 \begin{subfigure}{0.33\textwidth}\vspace{-1mm}
\centerline{\includegraphics[width=2.3in,height=1.7in]{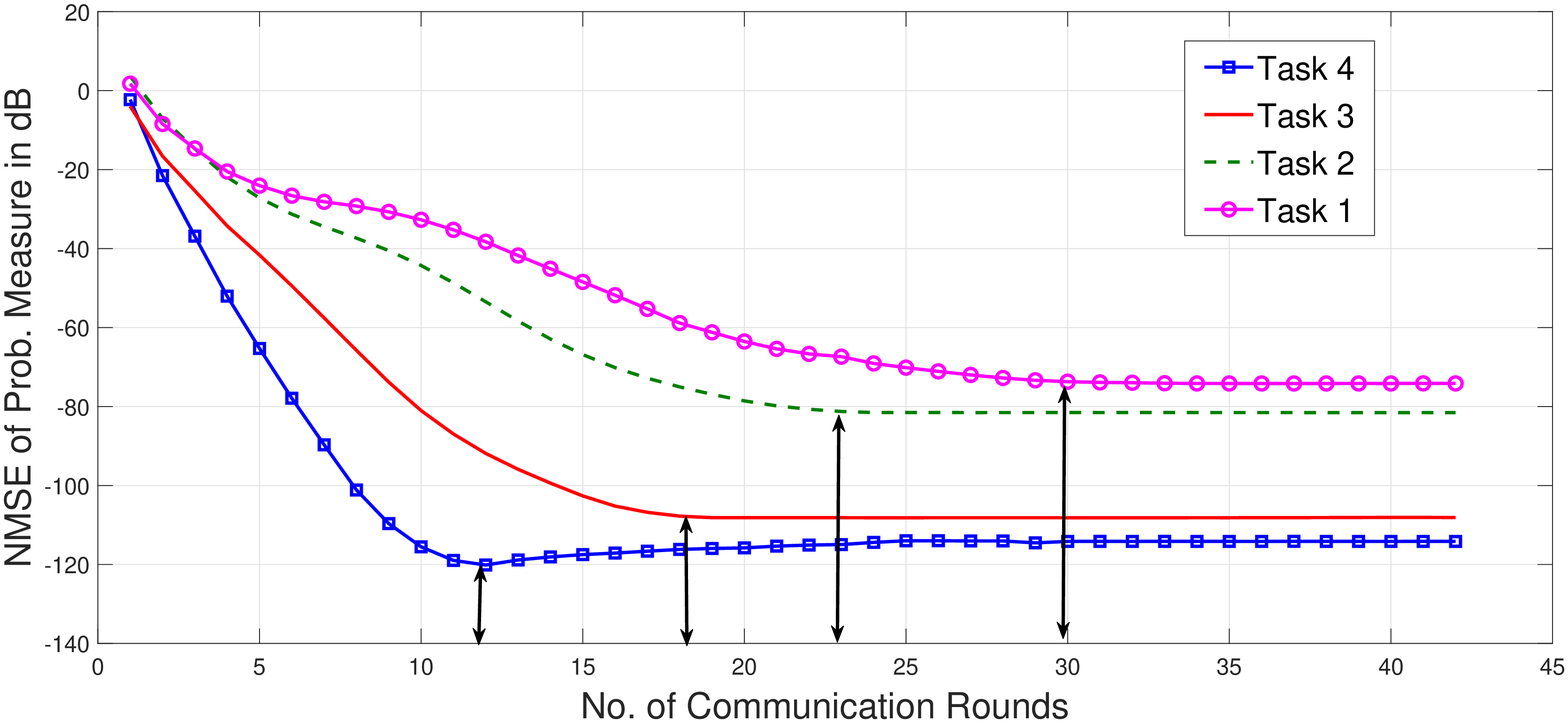}}\vspace{-1mm}
\caption{}
\label{EmLang_TaskAgnostic}\vspace{-1mm}
\end{subfigure}
\begin{subfigure}{.33\textwidth}\vspace{-1mm}
\centerline{\includegraphics[width=2.3in,height=1.7in]{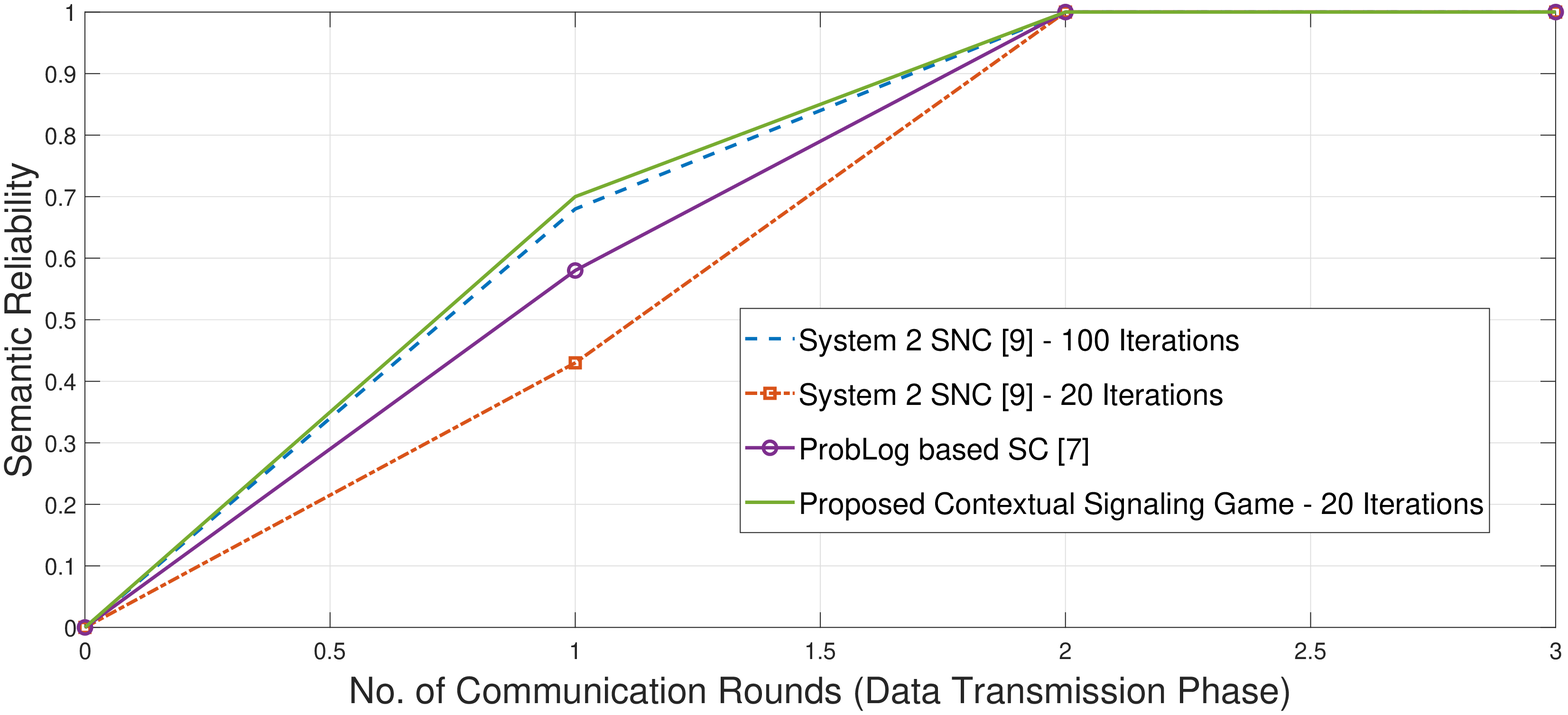}}\vspace{-1mm}
\caption{}
\label{EmLang_SemReliability}\vspace{-1mm}
\end{subfigure}
\begin{subfigure}{.33\textwidth}\vspace{-1mm}
\centerline{\includegraphics[width=2.4in,height=1.7in]{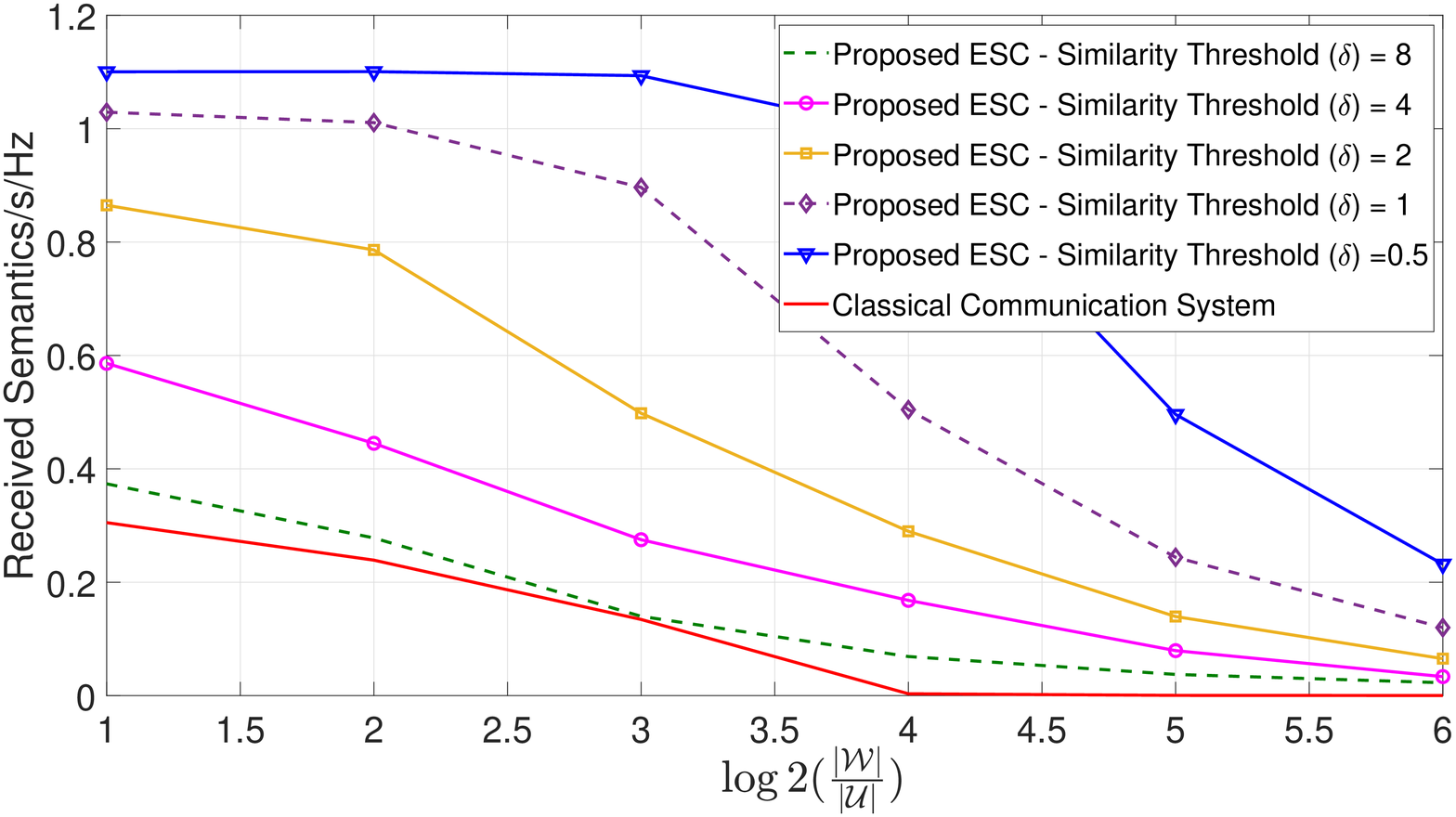}}\vspace{-1mm}
\caption{}
\label{EmLang_SemRate}\vspace{-1mm}
\vspace{-0mm}
\end{subfigure}\vspace{-1mm}
\caption{\scriptsize (a) Emergent language convergence for variations in tasks across time. This shows the decreasing language creation overhead. (b) Semantic reliability using the emergent language created. (c) Received semantics for different semantic similarity threshold (or semntic spaces), with $\abs{\mathcal{W}}=256$.}
\label{TaskAg_Relia_Semantics}
 \vspace{-3mm}\end{figure*}

Fig.~\ref{EmLang_TaskAgnostic} shows the task agnostic behavior of the emergent language constructed. Different tasks represent variations in the $\mathcal{W}$ that capture the event state descriptions. Task $1$ needs $30$ rounds to converge, while task $4$ needs only $10$ rounds. Hence, as time progresses the emergent language duration decreases, as illustrated in Fig.~\ref{Fig_timesplit} and a task agnostic language emerges between the nodes. In Fig.~\ref{EmLang_SemReliability}, we evaluate the semantic reliability using the proposed emergent language during data transmission phase. Our algorithm performs $75\%$ better than \cite{Mehdi2021} in terms of semantic reliability (for similar iterations). Fig.~\ref{EmLang_SemRate} shows the extracted semantics/s/Hz vs the decreasing $\abs{\mathcal{U}}$ (for a fixed $\abs{\mathcal{W}}$) at the listener using the metric defined in \eqref{eq_speakersemInfo}. The semantic rate decreases as $\abs{\mathcal{U}}$ decreases, but the point ($\abs{\mathcal{U}}$) at which the semantic rate starts a downward slope depends on the semantic threshold. This clearly shows that with increasing semantic similarity between causal states, the extracted semantic rate will be higher compared to wireless systems that do not exploit semantic awareness with respect to the listener.

 \vspace{-3mm}
\subsection{Performance Evaluation of the Proposed ESC with Causal Reasoning } \vspace{-1mm}

We followed the experimental setup of \cite{DeleuArxiv2022} for the data generation. In each
experiment, a random graph $G$ was generated from the Erd\"os-R\'enyi (ER) random graph model. The graph contained $d$ edges on average; the probability of creating an edge between two nodes was scaled accordingly. Given $G$, we uniformly assigned random edge weights to obtain a weight matrix $\bmW$. Edge weight $w_{ij}$ { (represent $\bmr_{ij}$ defined in Section~\ref{eq_CRM})} denotes the grounding for the corresponding relation between edges $i$ and $j$ in the graph. Given $\bmW$, we sampled $\bmX = \bmW^T \bmX + \bmz$, where $\bmz$ is a Gaussian noise, with zero mean and variance $=0.01$. The elements of $\bmW$ follows $\mathcal{N}(0,1)$. Once the full Bayesian Network is known, we used ancestral sampling to generate $100$ data points to fill our
data set $D$. The dataset generated here is quite realistic in that it can correspond to the various device measurements corresponding to an IoT environment,  aggregated at a central device with a radio transceiver (speaker). This generated data is fed to GFlowNet, which learns the causal 
structure forming the data $\bmX$. Instead of transmitting the entire piece of data, the speaker computes just the causal reasoning that gets encoded as $\bmu$ using the signaling game component and transmitted across the network, saving significant bits.  { The listener reconstructs the state description $\bmzh$ from the received signals $\bmy = \bmu + \sigma^2\bmn$, where $\bmn\sim \mN(0,1)$ and $\sigma^2$ corresponds to ans SNR of $20$~dB. $\bmu \in \mU$ will have unit amplitude and is chosen to be from $64-$QAM constellation set. The listener extracts the causal relations (and hence the DAG) from the state description using GFlowNet. Further, we define the following logical formula ``Is any neighboring nodes has sum of their energy greater than -5~dB". Given the result of this evaluation, whose result will be either $0$ or $1$, two different actions are taken respectively}. The simulations for the AI model training are all performed in google colab pro version that used an NVIDIA TESLA P100 GPU with a 16GB RAM and the code is implemented in \emph{PyTorch}. {The baselines considered here include the probabilistic logic SC system from \cite{SeoMehdiArxiv2022}, KLD based semantic baseline from \cite{Mehdi2021}, transformer based encoder and decoder \cite{XieTSP2021} and implicit semantic communication architecture \cite{XiaoICC2022}.} 

The training phase is performed using $25K$ different observations, and 100 mini-batches are considered. The GFlowNet weights are learned using this training phase. In the actual transmission phase (involving $10,000$ separate test data samples), the learned model is considered along with the emergent language in Algorithm~\ref{alg_ta_CSG} used for encoder/decoder. From an ESC perspective, measurements $\bmX$ represent the data received from $N$ ($=5$ here) different devices. At the listener, based on the quantized levels ($4$ possible levels considered) of the measurement vector decoded, different actions are taken. The learned graphical structure is represented by the adjacency matrix $\bmW$. Further, the encoded representation $\bmu_t$ is  sent across a BSC, with crossover probability $p$. 

\begin{figure*}[t]
 \begin{minipage}{.5\textwidth}\vspace{-3mm}
\centerline{\includegraphics[width=3.3in,height=1.6in]{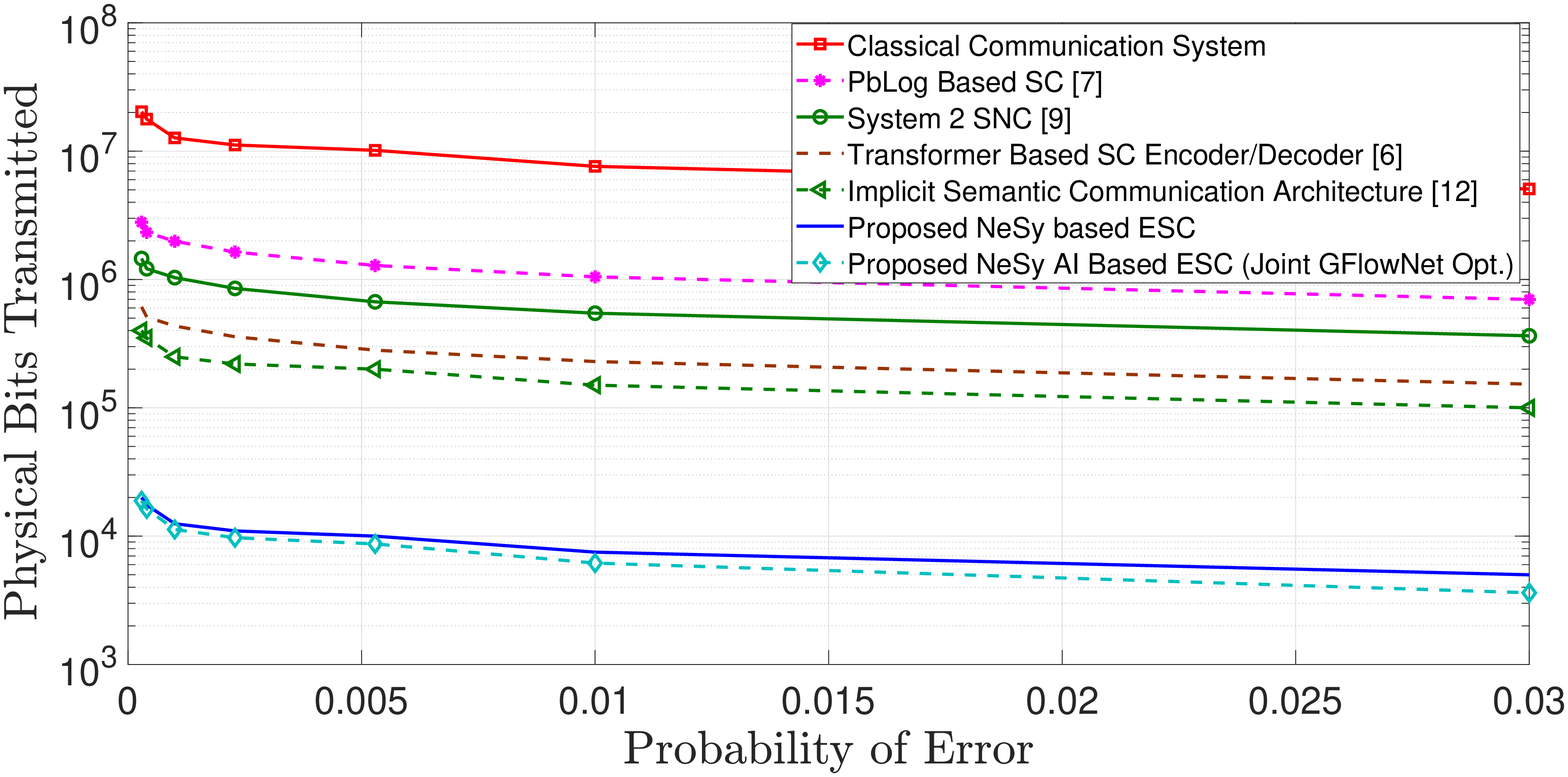}}\vspace{-1mm}
\caption{\scriptsize {Transmission Efficiency for Proposed ESC vs State of the Art.}}
\label{Semantics}\vspace{-0mm}
\end{minipage}
 \begin{minipage}{.5\textwidth}\vspace{-4mm}
\centerline{\includegraphics[width=3.3in,height=1.6in]{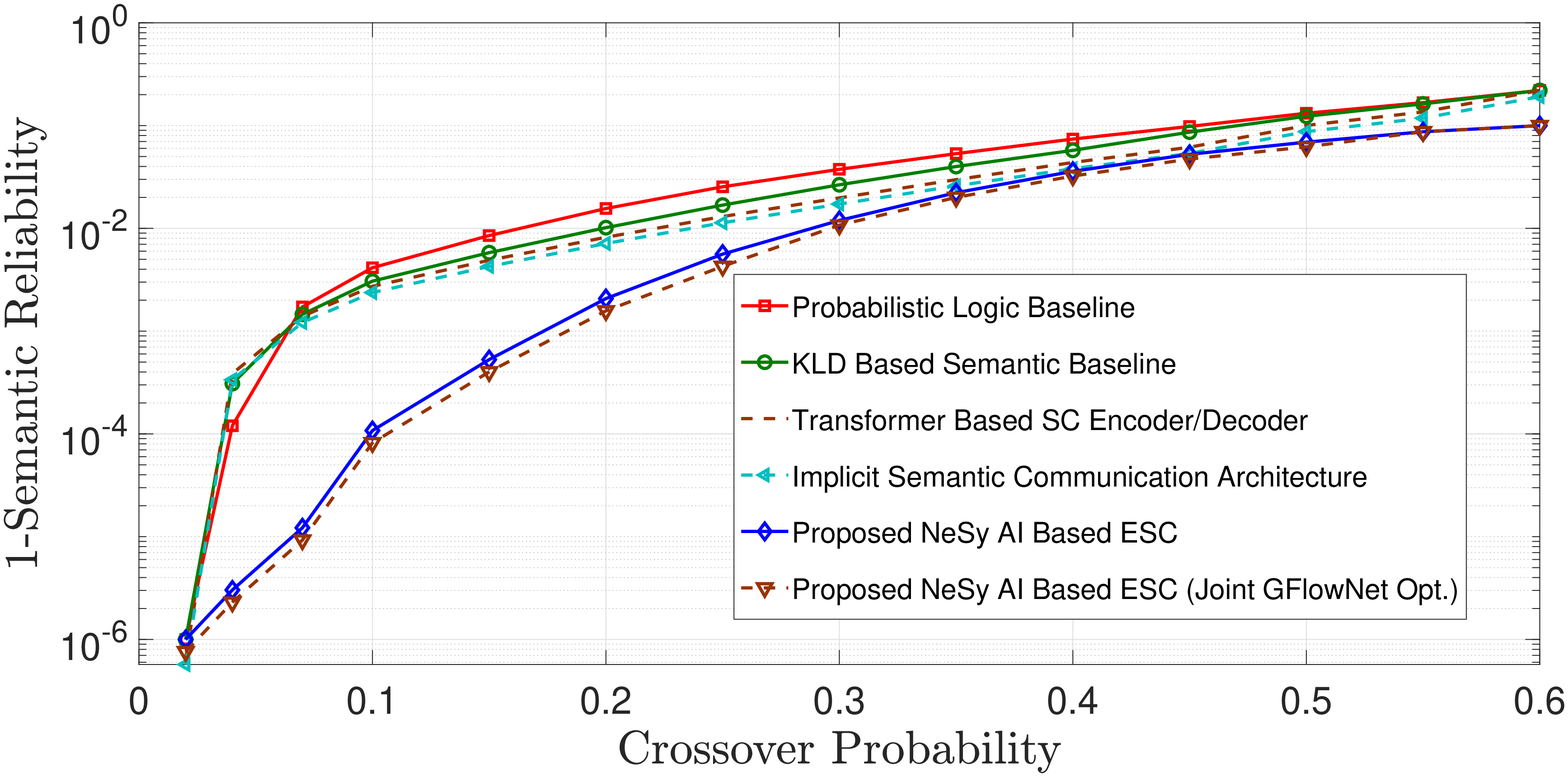}}\vspace{-1mm}
\vspace{-0mm}\caption{\scriptsize {Semantic Reliability for Proposed ESC vs State of the Art.}}
\label{NeSyvsPbLog}\vspace{-0mm}
\end{minipage}\vspace{-4mm}
\end{figure*}

Fig.~\ref{Semantics} shows that an ESC system based on NeSy AI can be more efficient (in terms of number of bits) to convey the same amount of semantic information compared to the conventional communication system {(represented by Legend ``Classical Communication System"). The Classical Communication System is implemented as a joint source channel coding scheme \cite{BourtsoulatzeTCCN2019} that aims to minimize the Euclidean distortion measure of reconstruction at the listener}. Specifically, we analyze the number of bits transmitted to get a particular probability of error (for $1,000$ events).  The number of bits transmitted is reduced by a factor of $1,000$ for the semantic system compared to classical system (without reasoning part) which explains the significance of our approach. Compared to state-of-the-art algorithms that do not consider causal reasoning, the proposed NeSy approach is more robust to bit errors. Our approach can still recover larger semantic information even with a probability of error of $3\%$. In Fig.~\ref{NeSyvsPbLog}, the semantic reliability of our ESC system shows significant improvements compared to state of the art methods. The evaluation is performed based on the semantic error probability (measured as $1-$ semantic reliability) vs $p$ for the BSC. The result shows that our ESC scheme has almost $95\%$ reduction in semantic error probability at $p=0.1$ compared to other methods, thanks to the causality and semantic awareness based encoding achieved using the game. We also show that the performance of our ESC approach in which the speaker and listener independently optimize the GFlowNet is very close to that with joint optimization of the GFlowNet parameters for both nodes. A joint optimization means that the listener would have communicated complete information about its actions and logical conclusions to the speaker, hence not preferred due to more overhead during training. As $p$ becomes closer to $0.5$, the semantic error probability converges to a higher value for all schemes since half of the bits get flipped at the listener.

  \vspace{-3mm}
\section{Conclusion}
 \vspace{-2mm}

In this paper, we have presented a new vision of SC systems entitled ESC that relies upon NeSy AI as a machinery to enable causal reasoning that allows it to be task agnostic over time with less training data. We have formulated a two-player contextual signaling game to learn an optimal representation space for transmission and extracting the semantic state description at the listener, which is termed the emergent language. The optimized local NE resulting from the AM updates ensures that the listener can recover the maximum of the relevant semantics with minimal transmission from the speaker. 
The causal reasoning component at the speaker and the symbolic logical reasoning at the listener are implemented using the recently emerged causal reasoning AI algorithm called GFlowNet. We have shown analytically that, by exploiting contextual information in emergent language, the average bit length for the semantic state description gets reduced compared to the classical communication system. Simulation results have demonstrated superiority of our proposed ESC in terms of improving the communication efficiency (minimal transmission) and reliability compared to classical communication and state of the art SC. 
\vspace{-1mm}\appendices
\vspace{-1mm}\section{\vspace{-1mm}Category Theory Preliminaries}
\label{app_catTheory}
\vspace{-1mm}
Please see the Appendix A of arxiv version of this paper \cite{ChristoTWCArxiv2022}.

\vspace{-2mm}\section{\vspace{-1mm}Proof of Lemma~\ref{lemma_semanticcat}}
\label{appendix_proof_lemma_semcat}
\vspace{-1mm}

It is straightforward to show that $\mL$ is a monoidal preorder, where the order is imposed by $\bmx \leq \bmy$, whenever there is a morphism from $\bmx$ to $\bmy$. For each object $\bmx$ in $\mL$, the representable copresheaf $h^\bmx = hom(\bmx,-)$ is given by the following conditional probability:.
    \vspace{-1mm}
\beq
    \vspace{-2mm}
\begin{array}{l}
    h^\bmx(\bmy) = \begin{cases}
    \pi(\bmy\mid x,c) \,\,\,& if \,\,\,\bmx \leq \bmy\\
    0,              & \text{otherwise}
\end{cases} 
    \end{array}
   \vspace{-1mm}
 \eeq
Assignment $\bmx\rightarrow h^\bmx$ can be shown to be an enriched functor as follows. An enriched functor $f=h^{\bmx}:\mL \rightarrow \mwL$ satisfies the property that $\mL(\bmx,\bmy) \leq \mwL(f(\bmx),f(\bmy))$. Here, we define $\mwL$ to be the category which contains all the enriched functors.  To show this, we need to first define the function $\mwL(f(\bmx),f(\bmy))$. 
\vspace{-1mm}
\beq
\vspace{-1mm}
\mwL(f(\bmx),f(\bmy)) =     \begin{cases}
1,  \,\,\, &if \,\, \bmx \leq \bmy \\
\frac{f(\bmx)}{f(\bmy)}, \,\,\, &if\,\, \bmx > \bmy
    \end{cases}
    \label{eq_morph_enrfunc}
\eeq
For $\bmx\leq \bmy$, $\mL(\bmx,\bmy) \leq 1 = \mwL(f(\bmx),f(\bmy))$. In the case, when $\bmx > \bmy$, $\mL(\bmx,\bmy) = 0 \leq \frac{f(\bmx)}{f(\bmy)},$ with $f(\bmx) \leq f(\bmy)$. Hence, $f=h^{\bmx}$ is an enriched functor. Hence, all the representable copresheaves $h^\bmx, \forall \bmx\in \mL$ within the context $c$ forms a [0,1] enriched category which we define as $\mwL$ (representing all possible worlds that follows from $\mL$). We denote the codomain of $h^\bmx$ as $\mH^\bmx$. Further, for completion, we can define the hom object between two representable copresheaves as follows, 
$\mwL(h^\bmx,h^\bmy) =    
\textstyle \sum\limits_{\bmd\in H^\bmy} h^\bmx(\bmd).
$
For any three objects $\bmx,\bmy,\bmz$, it satisfies $\mwL(h^{\bmx},h^{\bmz}) \!=\!\! \textstyle \sum\limits_{\bmd_1\in \mH^{\bmx}}\textstyle \sum\limits_{\bmd_2\in \mH^{\bmz}}\pi(\bmd_1\mid \bmx) \pi (\bmd_2 \mid \bmy) \!=\! \mwL(h^{\bmx},h^{\bmy})\mwL(h^{\bmy},h^{\bmz}).$
Hence, associative property of enriched categories is satisfied. $\mwL(h^{\bmx},h^{\bmx}) \!= \!1$ satisfies the identity morphism. Thus, $\mwL$ is a $[0,1]$-enriched category of representable copresheaves of the objects in $\mL$. 

\vspace{-1mm}\section{\vspace{-1mm}Proof of Proposition~\ref{Solution_NashEq}}
\vspace{-0mm}
\label{Proof_Solution_NashEq}
    Given a listener policy $\pi_{l,t}$, we optimize the speaker policy by maximizing the Lagrangian. 
\vspace{-0mm}\vspace{-1mm}\beq
\vspace{-0mm}\begin{aligned}
\pi_{s,t}^* &= \,\arg\max\limits_{\pi_{s,t}}F_s(\pi_{s,t},\pi_{l,t}), \,\, \mbox{where}, \\[-0mm] 
F_s(\pi_{s,t},\pi_{l,t}) &= \,- \mathbb{E}_{\bmu_t} 
 \left[S_s(\bmz_t;\bmu_t\mid \pi_{s,t},c,[\bmz_{t-1}])\right] \\ & - \lambda_s \mathbb{E} [V(\pi_{s,t},\pi_{l,t})] +  \sum\limits_{\bmz_t} \alpha_{\bmz_t}\left[\sum\limits_{\bmu_t}\pi(\bmu_t\mid \bmz_t)\right].
\end{aligned}\vspace{-1mm}
\eeq
It can be validated that $F_s(\pi_{s,t},\lambda_s)$ is {concave} in $\pi_{s,t}$ (follows from showing that term of the form $x\log \frac{x}{a}$ in \eqref{eq_speakersemInfo} is convex), hence the global maximum is obtained at the point where derivative of $F_s(\pi_{s,t},\pi_{l,t})$ is zero.
$\lambda_s$ represents the Lagrange multiplier for the inequality constraint associated with $V$. $\alpha_{\bmz_t}$ represents the Lagrange multiplier for the normalization of the distribution $\pi(\bmu_t\mid \bmz_t)$. $\balpha_s$ represents the vector of all $\alpha_{z_t}$.
Further, taking derivative w.r.t $\pi_{s,t}$, we obtain
\vspace{-1mm}\beq
\begin{array}{l}
 -\pi(\bmz_t) S_s(\bmz_t)\left(\log\pi_{s,t} - \log \pi(\bmu_t)+1\right) - \lambda_s \Big(c(\bmu_t) \\ - \log \pi_{l,t}\Big)  + \alpha_{\bmz_t} = 0, \implies \\ 
\pi_{s,t}   = \pi(\bmu_t)\\ \exp\left(\frac{-1}{S_s(\bmz_t\mid [\bmz_{t-1}])\pi(\bmz_t\mid c)}(\lambda_s (c(\bmu_t) - \log \pi_{l,t})-\alpha_{\bmz_t})-1\right).
\end{array}
\label{eq_speakTransmit}
\eeq
Similarly, for a fixed $\pi_{s,t}$, we can obtain $\pi_{l,t}$ by maximizing the Lagrangian \eqref{eq_pi_l_t}.
\begin{figure*}\vspace{-1mm}\beq
\begin{aligned}
\pi_{l,t}^* & = \arg\max_{\pi_{l,t}} F_l(\pi_{l,t},\pi_{s,t}), \\[-0mm] 
F_l(\pi_{l,t}^*,\pi_{s,t}) & = \sum\limits_{\bmu_t}\underbrace{\pi\left(\bmu_t\mid \bmz_t\right)\Big[\sum\limits_{\widehat{\bmz}_t} \pi(\widehat{\bmz}_t\mid \bmu_t)\log\frac{\pi(\widehat{\bmz}_t\mid \bmu_t)}{\pi(\widehat{\bmz}_t)}Z_{\widehat{z}_tz_t}\Big] S_s({\bmz}_t\mid c,[\bmz_{t-1}])}_{S_l(\widehat{\bmz}_t;\bmu_t\mid\pi_{s,t},\pi_{l,t},[\bmzh_{t-1}])}    -   \sum\limits_{\bmu_t}\alpha_{\bmu_t}\sum\limits_{\bmz_t} \pi_{l,t}.
\end{aligned}
\label{eq_pi_l_t}
\vspace{-1mm}
\eeq\vspace{-2mm}
\end{figure*}
\eqref{eq_pi_l_t} is non-concave since it is a summation of a convex and a concave function. Since a linear function is simultaneously convex and
concave, consider the first order Taylor series expansion of $\pi_{l,t} \log\pi_{l,t}$, using which we can rewrite $S_l(\widehat{\bmz}_t;\bmu_t\mid\pi_{s,t},\pi_{l,t},[\bmzh_{t-1}])$ as
\vspace{-1mm}\beq
\begin{aligned}
&\widehat{S}_l\left(\widehat{\bmz}_t;\bmu_t\mid \pi_{s,t},\pi_{l,t},[\bmzh_{t-1}]\right) \\ &= \pi\left(\bmu_t\mid \bmz_t\right)S_s({\bmz}_t\mid c,[\bmz_{t-1}])\sum\limits_{\widehat{\bmz}_t,\widehat{c}}\pi_{l,t}\big[1+\log\pi_{l,t-1} \\ & - \log\pi(\widehat{\bmz}_t)\big]Z_{\widehat{z}_tz_t}  + \mbox{const}. \nonumber
\label{eq_concave_Fl}
\end{aligned}
\vspace{-1mm}\eeq
Using the resulting approximate function obtained above which can be shown as a lower bound to the original $S_l(\widehat{\bmz}_t;\bmu_t\mid \pi_{s,t},\pi_{l,t},[\bmzh_{t-1}])$, we can solve the approximate Lagrangian function \eqref{eq_pi_l_t_approx}.
\begin{figure*}
    \vspace{-2mm}\beq
\begin{aligned}
\pi_{l,t}^* & = \arg\max_{\pi_{l,t}} \widehat{F}_l(\pi_{l,t},\pi_{s,t}), \\[-1mm] 
\widehat{F}_l(\pi_{l,t},\pi_{s,t}) & = \sum\limits_{\bmu_t}\widehat{S}_l\left(\widehat{\bmz}_t;\bmu_t\mid \pi_{s,t},\pi_{l,t},[\bmzh_{t-1}]\right)     -   \sum\limits_{\bmu_t}\alpha_{\bmu_t}\sum\limits_{\bmz_t} \pi(\widehat{\bmz}_t\mid \bmu_t) .
\end{aligned}
\label{eq_pi_l_t_approx}
\vspace{-2mm}
\eeq\vspace{-4mm}
\end{figure*}
We can now derive \eqref{eq_pi_l_t_der} w.r.t $\pi_{l,t}$
\vspace{-0mm}\beq
\begin{array}{l}
\pi\left(\bmu_t\mid \bmz_t\right)S_s({\bmz}_t\mid c,\left[\bmz_{t-1}\right])Z_{\widehat{z}_tz_t}\Big(\left(1+\log\pi_{l,t-1}\right) \\ - \log\pi(\widehat{\bmz}_t) - \frac{\pi_{l,t}}{\pi(\widehat{\bmz}_t)}\pi(\bmu_t)\Big) = \alpha_{\bmu_t} \implies\\[-0mm]
\pi_{l,t} \propto \frac{1}{\pi(\bmu_t)}\exp\left(\frac{-\alpha_{\bmu_t}}{\pi\left(\bmu_t\mid \bmz_t\right)Z_{\widehat{z}_tz_t}S_s({\bmz}_t\mid c,[\bmz_{t-1}])} +\log\pi_{l,t-1}\right)
\end{array}
\label{eq_pi_l_t_der}
\eeq
To arrive at $\pi_{l,t}$ above, we used the approximation $\pi(\bmzh_t) \approx \pi_{l,t}\pi(\bmu_t)$, since the probabilities are expected to be higher near the peak of $\pi_{l,t}$  and zero for all other $\pi(\bmzh_t\mid \bmu^{\prime}_t)$, with $\bmu^{\prime}_t\neq \bmu_t$. 
\vspace{-1mm}

\vspace{-1mm}
\section{\vspace{-1mm}Proof of Theorem~\ref{analysis_NashEq}}
\label{proof_analysis_NashEq}
\vspace{-1mm}

We consider that the cost of transmission is negligible, $c(\bmu) \!=\! 0$ and assume that the normalization constraints are satisfied with equality, hence $\balpha_{s}\!=\!0$. By using the inequality, $\sum_x p(x)\log q(x) \!\leq \!\sum_x p(x)\log p(x)$ (non-negativity of KLD), we can write (where $p, q$ are  $\pi_s, \pi_l$, respectively)
$F_s(\pi_{l},\pi_{s}) \geq \mathbb{E}\left[S_s(\bmz;\bmu\mid \pi_{s},c)\right]  - \lambda E\left[ \log \pi(\bmz\mid \bmu) \right].$
Substituting for $\pi(\bmz\mid \bmu)  $ using Bayes rule, we get
\vspace{-1mm}\beq
\vspace{-1mm}
\begin{aligned}
 &\mathbb{E}\left[S_s(\bmz;\bmu\mid \pi_s,c)\right] + \lambda_s \mathbb{E}\left[V(\pi_{s},\pi_{l})\right]   \\ &\geq  \mathbb{E}\left[S_s(\bmz;\bmu\mid \pi_s,c)\right] - \lambda_s \mathbb{E}\left[\log \frac{\pi(\bmu\mid \bmz)\pi(\bmz\mid c)\pi(c)}{\pi(\bmu)}\right]   
\\ & = (1-\lambda_s) \mathbb{E}\left[S_s(\bmz;\bmu\mid \pi_s,c)\right] + \lambda_s H_s(\bmz,c),
\end{aligned}
\vspace{-0mm}
\eeq
where $H_s(\bmz)$ is entropy of $\bmz$.
For $\lambda_s\! \in\! [0,1]$, the minimum is attained at $\mathbb{E}\left[S_s(\bmz;\bmu\!\mid\! \pi_s,c)\right]\! =\! 0$ and hence, 
$\mathbb{E}\left[S_s(\bmz;\bmu\!\mid\! \pi_s,c)\right] + \lambda_s \mathbb{E}[V(\pi_{s},\pi_{l})]  \geq   \lambda_s H_s(\bmz,c)$. 
It can be shown that the zero for $\mathbb{E}\left[S_s(\bmz;\bmu\!\mid\! \pi_s,c)\right]$ can be achieved when $\pi_s(\bmu\mid \bmz) \!=\! 1/K$, $\pi_l(\bmz\mid \bmu) \!= \!\pi(\bmz)$, and $H_s(\bmu)\! =\! \log K$, where $\mathcal{U}=K$. This implies that, the minimum can be achieved with a non-informative speaker when the constraint on the sender utility is weak. When $\lambda_s > 1$, the $F_s(\pi_{l},\pi_{s})$ is minimized when $\mathbb{E}\left[S_s(\bmz;\bmu\mid\pi_s,c)\right]$ is maximized. $\mathbb{E}\left[S_s(\bmz;\bmu\mid\pi_s,c)\right]$ is maximum when there is a bijection that preserves the semantics, that is $\psi:\mathcal{W}\rightarrow \mathcal{U}$ such that $\pi_s(\bmu\mid \bmz,c) \!=\! \delta_{\bmu,\psi(s)}, \pi_l(\bmz \mid \bmu,c) \!= \!\delta_{\bmu,\psi(s)}$. In this case, the listener can map accurately the received signal (assuming error-free communication) to the speaker intended semantics and thus extracting the maximum semantic information.

\vspace{-1mm}In the partial pooling case, where $\abs{\mW} > \abs{\mU}$, we first look at how the listener side extracts the semantic state. For simplicity of analysis, we assume that communication is error-free, which implies $\widehat{\bmu}=\bmu$. We can write 
$F_l(\pi_l,\pi_s) \leq \mathbb{E}_{\bmu_t}\left[S_l(\bmz;\bmu\mid \pi_s,\pi_l) \right]$, from which it is clear that, the maximum is attained  when there is a bijective mapping between $\bmz$ and $\bmu$, that contradicts our assumption for partial pooling case. Hence, the expression
\vspace{-1mm}\beq
\vspace{-1mm}\pi_l^* = \arg\max_{\pi_l}S_l(\bmzh;\bmu\mid \bmz,\pi_l),
\vspace{-1mm}\label{eq_maxSemInfo}
\eeq is maximized only when the overlap between the copresheaves $h^{\bmz}$ and $h^{\widehat{\bmz}}$ are maximized. Further, defining a semantic distance metric as $\norm{\bmz-\bmzh}_S = \sum\limits_{\bmY \in \mH^{\bmz} - (\mH^{\bmz} \cap \mH^{\bmzh})} S_s(\bmY; \bmz)$. Intuitively, it can be interpreted as a weighted ($S_s(\bmY;\bmz)$ are the weights) Hamming distance (HD) between the copresheaves represented by $\mH^{\bmz}$ and $\mH^{\bmzh}$. Hence, at equilibrium, the listener chooses the expression $\bmzh$ that is semantically closer to $\bmz$ or in other words, according to the optimization (equivalent to \eqref{eq_maxSemInfo}), $\bmzh^* = \arg \min_{\bmzh} \norm{\bmz-\bmzh}_S.$
Given that the listener chooses HD-based decoding, how should the speaker select the encoded signal to maximize the received semantic information (or minimize the HD)? The optimal encoded signal can be understood using a similar analysis as in \cite{JagerGEB2011}. The speaker can partition $\mathcal{W}$ into $\abs{\mathcal{U}}$ of them, with each denoted $\mathcal{W}_k$ and every state in $\mathcal{W}_k$ mapped to transmit signal $\bmu_k$. Let the listener's decoded state be $\bmzh_k$ corresponding to $\bmu_k$. The speaker wants the decoded expression
to be as close to the transmitted one as possible. Hence, the speaker will transmit $\bmu_k$ whenever the decoded state $\bmzh_k$ is semantically closest to all the states in $\mathcal{W}_k$. Such a partition of the $\mW$ is called a \emph{Voronoi tessellation} of $\mW$. 

\vspace{-2mm}\section{\vspace{-1mm}Proof of Theorem~\ref{theorem_se_representation}}
\label{appendix_proof_theorem_se_rep}
\vspace{-1mm}

 Assume that the length of each codeword $\bmu_i \in \mathcal{U}$ is $l_i$. Also, consider that the listener is able to disambiguate the ambiguous signaling strategy from the sender by inferring the context. Then the average semantic representation length for the vocabulary $\mathcal{U}$ can be lower bounded as (using the Kraft's inequality \cite{CoverThomas1991}) $\sum\limits_{\bmu_i \in \mathcal{U}}\pi(\bmu_i)l_i   \geq \sum\limits_{c\in \mathcal{C}}\pi( c) H(\bmz_i\mid c).
$
Moreover, we can upper bound the codeword length for the semantic representation as follows.
\vspace{-1mm}\beq
\vspace{-1mm}
\begin{array}{l}
\sum\limits_{\bmu_i \in \mathcal{U}}\pi(\bmu_i)l_i  \leq -\sum\limits_{c\in \mathcal{C}}\pi( c) \sum\limits_{\bmz_i}\pi(\bmz_i\mid c)\lceil\log \pi(\bmz_i\mid c)\rceil
\end{array}
\vspace{-0mm}
\eeq
For a causal reasoning system, where the contextual information is not taken into account $\!\!\sum\limits_{\bmu_i \in \mathcal{U}}\pi(\bmu_i)l_i  \geq H(\bmz_i)$,
and the upperbound is $\sum\limits_{\bmu_i \in \mathcal{U}}\pi(\bmu_i)l_i  \!\leq\!  \max\limits_{c\in \mathcal{C}}\sum\limits_{\bmz_i}\!\pi(\bmz_i\!\mid \!c)\lceil\log \pi(\bmz_i\!\mid\! c)\rceil.$ 
Further, we consider a classical system with $H_c(\bmz_i)$ representing the entropy. Classical systems are designed without a causal reasoning component and directly encodes all the entities. The bounds in this case are
{\vspace{-1mm}\beq
\vspace{-1mm}
\begin{array}{l}
\sum\limits_{\bmu_i \in \mathcal{U}}\pi(\bmu_i)l_i  \geq H_c(\bmz_i),\,\, \mbox{where,}\, H_c(\bmz_i) \stackrel{(a)}{>} H(\bmz_i), \,
\end{array}
\vspace{-1mm}
\eeq
 with, $H_c(\bmz_i)$ representing the entropy for classical wireless system. Statement (a) holds because the ESC system utilizes a uniform encoding for all states that convey equivalent meanings from the listener's standpoint (correspond to logical conclusions). In contrast, conventional wireless systems employ data statistics of $\bmz_i^0$ for encoding and do not account for the listener's perspective. This leads to a larger state space $\mW$ for conventional systems as opposed to ESC, which employs a reduced state space. Given that the conditional entropy obeys the relation $H(\bmz_i\mid c) \leq H(\bmz_i) < H_c(\bmz_i^0)$, the lower bound for semantic representation gets reduced further when we take into account the communication context, compared to a classical communication system.} Further, we look at the upper bound which is obtained as
\vspace{-1mm}\beq
\vspace{-1mm}
\begin{array}{l}
\sum\limits_{\bmu_i \in \mathcal{U}}\pi(\bmu_i)l_i  \leq  \max\limits_{c\in \mathcal{C}}\sum\limits_{\bmz_i}\pi(\bmz_i\mid c)\left[\sum\limits_{\bms_i \in \bmz_i}\lceil\log \pi(\bms_i\mid c)\rceil\right].
\end{array}
\label{eq_ub_sr_classical}
\vspace{-0mm}
\eeq
The upper bound \eqref{eq_ub_sr_classical} follows from the reasoning that the classical system should be designed considering the worst-case scenario such that the listener can decode properly with an arbitrarily low probability of error. This results in a higher upper bound for classical system compared to ESC.

\vspace{-2mm}\section{\vspace{-1mm}Proof of Theorem~\ref{theorem_se_error}}
\label{proof_theorem_se_error}
\vspace{-1mm}

In our envisioned ESC model, we have the following set of inequality, that mimics the data processing inequality in information theory \cite{CoverThomas1991}, 
$S_s(\bPhi;\bmz) \geq S_s(\bPhi;\bmu) \geq S_s(\bPhi;\bmzh)$.
Here $\bPhi$ (logical formulas) represents the semantic category space represented by $\bmz$. This follows from the Markov Chain $\bPhi \rightarrow \bmz \rightarrow \bmu \rightarrow \bmuh \rightarrow \bmzh \rightarrow \widehat{\bPhi}$. In a classical system, the state description represented as $\bmz_c$ does not consider the hidden semantics and may contain irrelevant information. In such a case, where the semantic category space is excluded, the corresponding Markov Chain would be $ \bmz_c \rightarrow \bmu \rightarrow \bmuh \rightarrow \bmzh_c$. Moreover, $H(\bmz_c\mid\bmzh_c)\geq H(\bmz_c\mid\bmu)\geq H(\bmz\mid\bmu)$, assuming the same representation space for ESC and classical systems. Hence, we have the following Fano's inequality \cite{CoverThomas1991}, for a classical system, 
$P_e = p(\bmzh_c \neq \bmz_c) \geq \frac{H(\bmzh_c\mid \bmz_c) - 1}{\log \abs{\mathcal{W}}}$.
Next, we look at how the semantic probability error measure differs from the classical error quantity above. We define $\bme$ as the random variable that captures the error in semantics represented by $\bmz$ and $\bmzh$. $\bme$ denotes the set $\mH^{\bmz} - (\mH^{\bmz} \cap \mH^{\bmzh})$, that represents the semantic space that cannot be represented by $\bmzh$. Moreover, consider $H(\bme,\bPhi\mid \bmzh) = H(\bPhi\mid \bmzh) + H(\bme\mid \bPhi, \bmzh) = H(\bPhi\mid \bmzh) = H(\bmz\mid \bmzh)$. Further, we proceed following similar derivation as for Fano's inequality. Rewriting, $H(\bme,\bPhi\mid \bmzh) \!=\! H(\bme\mid \bmzh) \!+ \!H(\bPhi\mid \bme, \bmzh)$. With probability $p(\mH^{\bmz} = \mH^{\bmzh})$, we have $H(\bPhi\mid \bme, \bmzh) = 0$. With probability  $S_e = 1-p(\mH^{\bmz} = \mH^{\bmzh})$, we have $H(\bPhi\mid \bme, \bmzh) = H(\bmz\mid \bme, \bmzh) \leq H(\bmz) \leq \log \abs{\Omega} $. Further, we are able to conclude as
\vspace{-1mm}\beq\vspace{-1mm}
\begin{array}{l}
H(\bmz\mid \bmzh) \leq H(\bme\mid \bmzh) + S_e \log \abs{\Omega} \\ \implies 
S_e \geq \frac{H(\bmz\mid \bmzh) - H(\bme\mid \bmzh)}{\log \abs{\mathcal{W}}},
\end{array}
\vspace{-0mm}\eeq
where $H(\bme) \geq 1$, with equal to happens only when the intersection of the semantic space represented by $\bmz$ and $\bmzh$ is a null set. Moreover, $H(\bmz_c\mid\bmzh_c) \geq H(\bmz\mid\bmzh)$. Combining these two, we can conclude that even with some bit errors in the physical transmission, the semantics represented by $\bmzh$ can be similar (with its semantic space having significant intersection to $\mH^{\bmz}$). 

\vspace{-5mm}
\bibliographystyle{IEEEbib}
\bibliography{semantic_refs}
\vspace{-8mm}
 
\vspace{-6mm}

\vfill

\end{document}

%% file: defines.tex
\newcommand{\beq}{\begin{equation}}
\newcommand{\eeq}{\end{equation}}

\newcommand{\mP}{\mbox{$\mathcal P$}}

\newcommand{\bmn}{\boldsymbol{n}}

\DeclareMathOperator*{\argmax}{arg\,max}
\DeclareMathOperator*{\argmin}{arg\,min}

\newcommand{\mS}{\mathcal{S}}
\newcommand{\mI}{\mathcal{I}}
\newcommand{\mO}{\mathcal{O}}

\newcommand{\mN}{\mathcal{N}}

\newcommand{\mR}{\mathcal{R}}

\newcommand{\mW}{\mathcal{W}}
\newcommand{\mZ}{\mathcal{Z}}

\newcommand{\mX}{{\mathcal X}}

\newcommand{\mU}{{\mathcal U}}

\newcommand{\bmd}
{\boldsymbol{d}}

\newcommand{\dsum}{\displaystyle\sum}

\def\adots{\mathinner{\mskip0mu\raise0pt\vbox{\kern7pt\hbox{.}}\mskip3mu
          \raise4pt\hbox{.}\mskip3mu\raise8pt\hbox{.}\mskip0mu}}

\newcommand{\bmc}{{\boldsymbol c}}

\newcommand{\bmW}{{\boldsymbol W}}

\newcommand{\mH}{\mathcal{H}}

\usepackage{bm}

\newcommand{\bme}{{\bm e}}
\newcommand{\bmx}{{\bm x}}
\newcommand{\bmy}{{\bm y}}
\newcommand{\bmv}{{\bm v}}

\newcommand{\bmw}{{\bm w}}

\newcommand{\bmq}{{\bm q}}
\newcommand{\bmu}{{\bm u}}

\newcommand{\bmz}{\bm z}
\newcommand{\bmzh}{\widehat{\bmz}}

\newcommand{\bms}{{\bm s}}

\newcommand{\bmA}{{\bm A}}
\newcommand{\bmr}{{\bm r}}
\newcommand{\bmM}{{\bm M}}

\newcommand{\bma}{{\bm a}}

\newcommand{\btheta}{\boldsymbol{\theta}}
\newcommand{\balpha}{\boldsymbol{\alpha}}

\newcommand{\bphi}{\boldsymbol{\phi}}
\newcommand{\bmeta}{\boldsymbol{\eta}}